\documentclass[times,review]{elsarticle}

\usepackage{amssymb}
\usepackage{amsmath}
\usepackage[table]{xcolor}
\usepackage{booktabs}
\usepackage{multirow}
\usepackage{makecell}
\usepackage{caption}
\usepackage{pifont}
\usepackage{tabularx}
\usepackage{wrapfig}
\newcommand{\cmark}{\ding{51}}
\newcommand{\xmark}{\ding{55}}
\definecolor{lightgreen}{RGB}{232,246,235}
\definecolor{mygreen}{RGB}{0,128,0}
\usepackage{geometry}
\journal{Pattern Recognition}

\begin{document}

\begin{frontmatter}

\title{GaussFusion: Towards Multimodal 3D Gaussian Pretraining}

\author[inst1]{Zhixuan You\corref{cor1}}
\ead{2223525386@stu.xjtu.edu.cn}

\author[inst1]{Jihua Zhu\corref{cor1}}
\ead{zhujh@xjtu.edu.cn}

\author[inst1]{Yiding Sun\corref{cor1}}
\ead{sunyiding@stu.xjtu.edu.cn}

\author[inst1]{Zihao Guo}
\ead{1991002470@stu.xjtu.edu.cn}

\author[inst1]{Haozhe Cheng}
\ead{chz97@stu.xjtu.edu.cn}

\author[inst1]{Dongxu Zhang}
\ead{zhangdongxu@stu.xjtu.edu.cn}

\author[inst1]{Lin Chen}
\ead{chenlin21@hnu.edu.cn}

\author[inst2]{Hainan Luo}
\ead{luohainan@hitrobot.com.cn}

\cortext[cor1]{Corresponding author}

\affiliation[inst1]{organization={School of Software Engineering},
            addressline={Xi'an Jiaotong University},
            city={Xi'an},
            postcode={710100},
            state={Shaanxi},
            country={China}}

\affiliation[inst2]{organization={Wuhu HIT Robot Technology Research Institute Co., Ltd.},
            city={Wuhu},
            state={Anhui},
            country={China}}

\begin{abstract}
3D Gaussian Splatting provides an explicit representation that jointly models geometry and appearance, serving as a scalable foundation for 3D representation learning. Existing pre-training methods for Gaussian representations, such as masked Gaussian reconstruction, primarily capture local structures but offer limited semantic supervision.
In this paper, we propose GaussFusion, a multimodal pre-training framework for 3D Gaussian representations. GaussFusion integrates image and text supervision into masked Gaussian modeling through cross-modal semantic alignment, enabling the Gaussian encoder to learn both visual and language-level semantic information during pre-training.
To better adapt masked modeling to the non-uniform distribution of Gaussian primitives, we further propose Gaussian Salience-guided Multi-scale Hole Masking (GSHM). GSHM constructs spatially continuous masked regions based on Gaussian salience. By applying hole masks at multiple scales, GSHM encourages the encoder to capture both fine-grained local patterns and broader structural dependencies.
Extensive experiments on downstream tasks demonstrate that GaussFusion improves the transferability of Gaussian representations. Notably, GaussFusion outperforms Gaussian-MAE on ModelNet40 and ScanObjectNN (PB-T50-RS) by 0.61\% and 3.85\%, respectively.
\end{abstract}


\end{frontmatter}
\section{Introduction}\label{sec:introduction}
3D Gaussian Splatting (3DGS)\cite{kerbl2023gaussian} represents 3D objects and scenes with anisotropic Gaussian primitives. Unlike point clouds, which typically encode discrete geometric samples, 3DGS assigns opacity, scale, rotation, and color to each primitive, thereby modeling geometry and appearance within a unified explicit representation. With high-quality rendering capability and efficient optimization, 3DGS has become an important paradigm for 3D visual representation. However, most existing studies use 3DGS primarily for reconstruction and rendering, leaving the structural and semantic information in the Gaussian parameter space insufficiently explored for 3D understanding.

This gap is important for 3D representation learning, where transferable models are still limited by the scarcity of large-scale, high-quality annotated 3D data. Although PointNet\cite{qi2017pointnet} and PointNet++\cite{qi2017pointnetplusplus} show that deep neural networks can learn effective features directly from point clouds, scaling point cloud pre-training remains difficult because point cloud datasets require costly acquisition, cleaning, and annotation. In contrast, 3DGS data can be automatically generated from existing 3D models, multi-view images, and rendering pipelines\cite{schonberger2016sfm,mildenhall2020nerf,muller2022instant}, offering greater scalability for self-supervised learning. Therefore, learning structural and appearance priors from large-scale 3DGS data and transferring the learned representations to downstream tasks, such as point cloud classification, segmentation, and few-shot recognition, provides a feasible way to reduce the dependence on annotated 3D data.

\begin{figure}
	\centering
	\includegraphics[width=\linewidth]{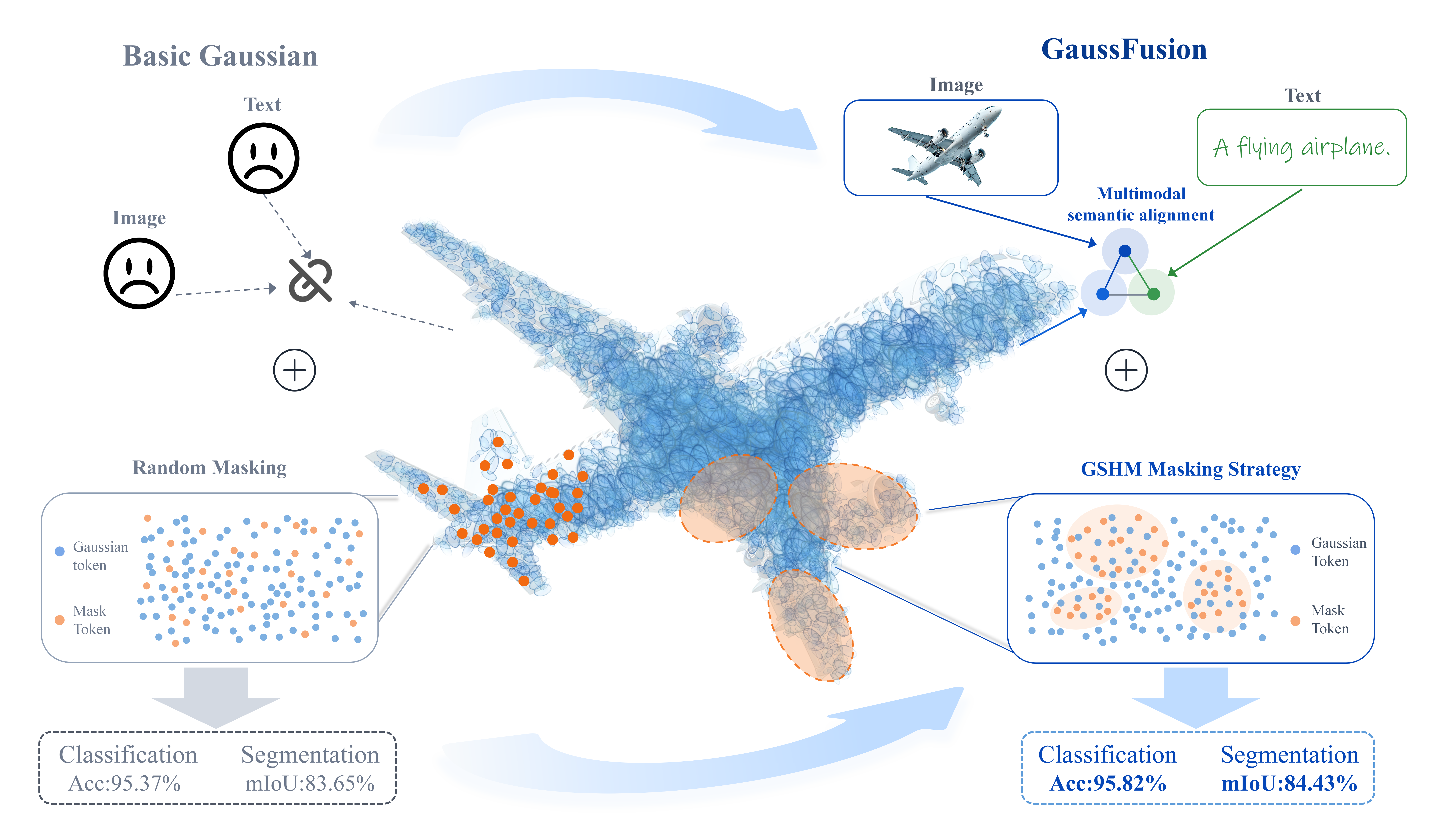}
	\caption{\textbf{Motivation and overview of GaussFusion.} Compared with basic masked Gaussian modeling, GaussFusion introduces image-text semantic alignment and GSHM masking to select salient, spatially coherent Gaussian regions, improving local structure learning and semantic transferability.}\label{fig1}
\end{figure}

Existing 3D pre-training methods are mainly designed for point clouds and can generally be divided into generative and contrastive approaches. Generative methods, such as Point-BERT\cite{yu2022pointbert} and Point-MAE\cite{pang2022pointmae}, learn contextual 3D information through masked point patch modeling or masked autoencoding reconstruction. Contrastive and cross-modal methods further introduce supervision from 2D images, text, and pre-trained vision models to enhance point cloud representations. For example, CrossPoint\cite{afham2022crosspoint}, ACT\cite{dong2023act}, and ReCon\cite{qi2023recon} leverage cross-modal information to improve the semantic expressiveness of point cloud features. These studies show that mature 2D vision-language models can provide effective semantic priors for 3D representation learning. Recently, several studies have started to perform pre-training directly on 3DGS representations. Gaussian-MAE, proposed in ShapeSplat\cite{ma2025shapesplat}, introduces the masked autoencoder paradigm into Gaussian primitive modeling. By randomly masking and reconstructing Gaussian attributes, it enables self-supervised learning of Gaussian representations and achieves strong transfer performance on multiple downstream point cloud tasks. SceneSplat\cite{li2025scenesplat} further extends this direction to scene-level 3D representation learning and improves global modeling capability with large-scale scene data.

Existing 3DGS-oriented pre-training methods have shown that Gaussian representations can be effectively transferred to downstream 3D tasks. However, most of these methods are still driven by masked Gaussian reconstruction alone. This objective helps the model recover local geometry and primitive attributes, but it provides limited guidance for learning semantic distinctions across object categories and scenes. When objects share similar shapes or local structures, or when observations are heavily occluded, reconstruction based only on Gaussian parameters may fail to learn sufficiently discriminative representations. Meanwhile, pre-trained image and vision-language models\cite{radford2021clip,dosovitskiy2021vit} have shown strong generalization capability in visual semantic modeling. This motivates the integration of image appearance cues, textual semantics, and 3DGS parameters, so that Gaussian representations can learn both local structural details and global semantic information for more transferable 3D representation learning.

Beyond the design of the training objective, multimodal Gaussian pre-training also requires a masking strategy that is compatible with cross-modal semantic alignment. In multimodal pre-training, image and text features can provide semantic supervision for Gaussian representations. However, such supervision cannot be fully exploited if the masked Gaussian regions are selected randomly. Unlike uniformly sampled points, Gaussian primitives are irregularly distributed and exhibit large variations in opacity, scale, and spatial coverage. Random masking may either remove many low-information primitives that contribute little to semantic learning or disrupt a small number of visually salient primitives that correspond to key object parts. As a result, the reconstruction targets may be weakly associated with the semantic cues provided by images and text, leading to large variations in reconstruction difficulty and limiting the effective use of multimodal supervision. These challenges indicate that effective multimodal Gaussian pre-training requires a masking mechanism that selects visually salient and structurally informative regions while preserving spatial coherence across scales.

To address these limitations, this paper proposes GaussFusion, a multimodal self-supervised pre-training framework for 3D Gaussian representations. Gaussian primitives are used as the fundamental input representation. Local grouping and masked reconstruction are employed to learn 3D structural and attribute information, while image and text encoders are introduced to provide external semantic supervision, enabling Gaussian representations to remain aligned with visual and linguistic features during pre-training. Furthermore, a salience-guided multi-scale hole masking strategy, termed Gaussian Salience-guided Multi-scale Hole Masking (GSHM), is proposed. This strategy estimates the importance of local regions according to Gaussian attributes, such as opacity and scale, and constructs spatially meaningful masked regions, thereby encouraging the model to focus on Gaussian regions that are more critical for semantic and structural understanding during reconstruction. Through the joint optimization of reconstruction learning and cross-modal alignment, GaussFusion learns 3D representations that preserve geometric structure, encode appearance information, and support semantic discrimination, as shown in Fig.~\ref{fig1}.

The main contributions of this paper are summarized as follows:
\begin{itemize}
	\item We identify that existing Gaussian pre-training methods mainly rely on reconstructing masked Gaussian attributes, which limits their ability to capture high-level semantic information for 3D understanding. To address this problem, we propose GaussFusion, a multimodal self-supervised pre-training framework that integrates images, text, and 3D Gaussian parameters for semantic-aware Gaussian representation learning.
	\item To improve the compatibility between masked Gaussian modeling and multimodal semantic alignment, we propose Gaussian Salience-guided Multi-scale Hole Masking (GSHM). GSHM adapts the masking process to the non-uniform distribution and structural importance of Gaussian primitives by constructing spatially continuous masks over salient regions at multiple scales.
	\item Extensive experiments on multiple downstream point cloud tasks demonstrate that GaussFusion improves the transferability and robustness of Gaussian representations under challenging and data-limited scenarios. These results suggest that 3DGS can serve not only as a rendering-oriented representation, but also as a scalable representation space for multimodal 3D pre-training and general 3D understanding.
\end{itemize}
\section{Related Work}\label{sec:related-work}

\subsection{Point Cloud Representation Learning}
Point cloud representation learning aims to extract discriminative features from unordered 3D point sets to support downstream tasks such as classification, segmentation, and retrieval~\cite{wang2026pointrft,zhang2026cmhanet}. Early studies mainly follow the supervised learning paradigm. PointNet\cite{qi2017pointnet} first enables end-to-end modeling of point sets and addresses the permutation invariance of point clouds through symmetric functions. Based on this foundation, PointNet++\cite{qi2017pointnetplusplus} further introduces hierarchical local region modeling to enhance the perception of local geometric structures. These methods mark a shift from global point-set aggregation toward hierarchical local structure modeling in point cloud representation learning. However, their performance gains usually rely on large-scale and high-quality annotated data, while the high annotation cost of 3D data limits their further deployment in practical scenarios.

To alleviate the shortage of annotated data, self-supervised point cloud pre-training has become an important direction in 3D representation learning\cite{prcv,sun2026align,SUN2026112800,li2025pointdico}. Point-BERT\cite{yu2022pointbert} divides point clouds into local point patches and learns contextual representations through masked point modeling. Point-MAE\cite{pang2022pointmae} adopts a masked autoencoder framework and directly reconstructs the masked local point patches. Point-M2AE\cite{zhang2022pointm2ae} further introduces multi-scale structures to strengthen hierarchical geometric modeling. Recent studies have also explored teacher-guided masking and contrastive patch-graph learning to improve the semantic quality of point cloud representations\cite{cheng2024ptm,zhou2026cpg}. Mamba-based and parameter-efficient adaptation methods further improve the efficiency of point cloud foundation models, as shown by PointMamba\cite{liang2024pointmamba} and Mantis\cite{guo2026mantis}. These studies show that masked reconstruction-based pre-training and efficient sequence modeling provide effective supervision signals for 3D understanding tasks. In contrast, this work performs pre-training on 3DGS data, which is more scalable and contains richer attributes, and then transfers the learned structural and appearance priors to point cloud understanding tasks.

\subsection{Gaussian Representation Learning}
3D Gaussian Splatting\cite{kerbl2023gaussian} represents 3D objects or scenes as a set of 3D Gaussian primitives and achieves high-quality real-time rendering through differentiable rasterization. Early studies on 3DGS mainly focus on reconstruction quality, rendering efficiency, and geometric consistency. For example, SuGaR\cite{guedon2024sugar} improves the alignment between Gaussians and object surfaces through surface alignment constraints and supports mesh extraction. Compressed 3D Gaussian Splatting\cite{lee2024compact} reduces the storage cost of Gaussian representations through compression and quantization strategies.

As the application scope of 3DGS continues to expand, recent studies have increasingly explored its capacity for semantic representation. Feature 3DGS\cite{zhou2024feature} distills features from 2D foundation models into Gaussian representations, enabling Gaussian fields to carry semantic features. LangSplat\cite{qin2024langsplat} embeds language features into 3DGS for open-vocabulary 3D queries. Furthermore, ShapeSplat\cite{ma2025shapesplat} constructs a large-scale object-level Gaussian dataset and proposes Gaussian-MAE, which learns self-supervised representations by reconstructing masked Gaussian attributes. SceneSplat\cite{li2025scenesplat} extends Gaussian representation learning to scene-level understanding and incorporates vision-language pre-training to enhance semantic modeling. These studies show that 3DGS is not only an explicit representation for rendering, but also a learnable representation for 3D understanding. However, existing object-level Gaussian pre-training still mainly relies on a single reconstruction objective, and dedicated masking mechanisms for the non-uniform distribution of Gaussian primitives remain underexplored.

\subsection{Multimodal-Guided 3D Representation Methods}
The development of 2D vision-language models\cite{radford2021clip} provides new forms of supervision for 3D representation learning. Compared with 3D annotated data, 2D images and text are easier to obtain, and large-scale vision-language models have learned strong appearance and semantic priors. Therefore, many studies attempt to enhance point cloud representations with 2D or language information\cite{zhu2023pointclipv2}. Joint representation learning has been used to connect text and point clouds through an intermediate image space\cite{huang2024joint}, while cross-modal knowledge transfer provides complementary guidance for point cloud representation learning\cite{zhang2025crossmodal}. CrossPoint\cite{afham2022crosspoint} improves the discriminative ability of point cloud features through cross-modal contrastive learning between point clouds and rendered images. Related methods introduce cross-modal information bottlenecks or hyperbolic contrastive objectives to strengthen semantic correspondence between images and point clouds\cite{cheng2024multitrusted,hu2024hyperbolic}. ACT\cite{dong2023act} uses pre-trained 2D models as teachers to guide 3D networks toward more semantic feature representations. ReCon\cite{qi2023recon} further combines reconstruction learning with cross-modal contrastive learning, allowing local geometry recovery and global semantic discrimination to complement each other. ULIP\cite{xue2023ulip} maps language, images, and point clouds into a unified representation space, improving open-vocabulary semantic understanding in point cloud models. However, most existing methods take point clouds as the 3D input and mainly focus on aligning point cloud features with image or text features. Research that directly introduces image and text supervision into 3DGS pre-training to improve Gaussian representation learning remains relatively limited.
\begin{figure}
	\rmfamily
	\centering
	\includegraphics[width=\linewidth]{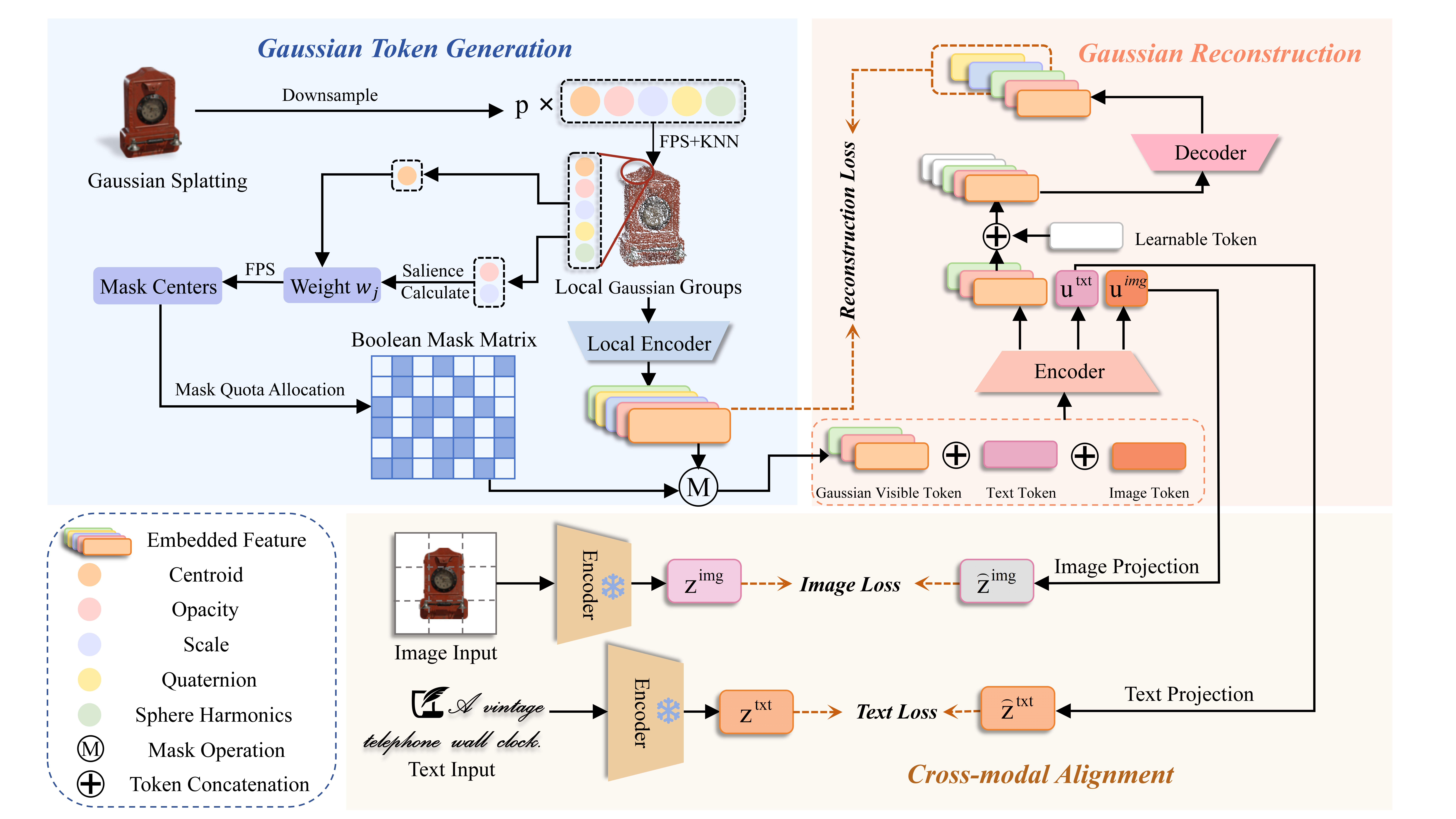}
	\caption{\textbf{Overall framework of GaussFusion.} GaussFusion generates Gaussian tokens from 3D splats and applies Gaussian Salience-guided Multi-scale Hole Masking (GSHM) to local Gaussian groups. Visible Gaussian tokens are combined with learnable image and text tokens in a multimodal encoder. The decoder reconstructs masked Gaussian groups, while frozen image and text encoders provide cross-modal supervision for learning transferable 3D representations.}
	\label{fig:framework}
\end{figure}

\section{GaussFusion}\label{sec:method}
The goal is to learn transferable representations from 3D Gaussian primitives. To this end, a pre-training framework is constructed by jointly integrating local Gaussian structure modeling, masked reconstruction learning, and cross-modal semantic alignment. The overall framework of the proposed GaussFusion is shown in Fig.~\ref{fig:framework}. The input 3D Gaussian primitives are first divided into local Gaussian groups and encoded as tokens. Then, the GSHM strategy is applied to mask a subset of Gaussian regions. During the reconstruction of the masked geometric and attribute information, image and text features are introduced for cross-modal semantic alignment.

\textbf{Gaussian Token Generation}
Following Gaussian-MAE\cite{ma2025shapesplat}, each Gaussian primitive is treated as an attributed point and is formally analogous to a point cloud input. Given a scene composed of $N$ Gaussian primitives, the input is denoted as
\begin{equation}
	\mathcal{G} = \{g_i\}_{i=1}^{N}, \quad
	g_i = (\mu_i, \alpha_i, s_i, q_i, SH_i) \in \mathbb{R}^{59},
\end{equation}
where $\mu_i \in \mathbb{R}^{3}$ denotes the position of the Gaussian center, $\alpha_i \in \mathbb{R}^{1}$ denotes opacity, $s_i = (s_{i,x}, s_{i,y}, s_{i,z}) \in \mathbb{R}^{3}$ denotes the scale along the principal axes, $q_i \in \mathbb{R}^{4}$ denotes the rotation parameter in quaternion form, and $SH_i \in \mathbb{R}^{48}$ denotes the spherical harmonics coefficients.

GaussFusion first organizes the unordered Gaussian set $\mathcal{G}$ into local structural units $\mathcal{N}_j$. Gaussian group centers are selected by farthest point sampling over Gaussian attributes, denoted as $c_j = \operatorname{FPS}(\mathcal{G})$. Then, according to the selected attribute indices, a local neighborhood is constructed around each center in the attribute space as
$\mathcal{N}_j = \operatorname{KNN}(\mathcal{G}_{\mathrm{attr}}, c_{j,\mathrm{attr}})$.
The Gaussian attributes in each local neighborhood are then fed into a geometric encoder to obtain the Gaussian token:
\begin{equation}
	z_j = E_g(\mathcal{N}_j) + P(c_j),
\end{equation}
where $E_g(\cdot)$ denotes the Gaussian local encoder and $P(\cdot)$ denotes the positional embedding function.

\textbf{Multimodal Encoding} GaussFusion applies the GSHM masking strategy to Gaussian tokens and feeds only the unmasked visible tokens into the Transformer encoder\cite{vaswani2017attention}. Meanwhile, image tokens and text tokens are inserted at the beginning of the sequence to receive semantic supervision from external modalities. Image and text semantics are therefore not used as auxiliary information in a post-processing stage, but directly participate in the formation of 3D representations during pre-training, as detailed in Section~\ref{sec:3.2}.

\textbf{Cross-Modal Alignment and Reconstruction} The reconstruction branch constrains the model to infer missing Gaussian parameters from local context. The cross-modal branch aligns Gaussian representations with semantic features produced by frozen image and text encoders, enabling 3D representations to absorb view-level appearance information and category-level semantic information. The two branches are jointly optimized within the same Gaussian encoding space.

\subsection{Gaussian Salience-guided Multi-scale Hole Masking}
\begin{wrapfigure}{r}{0.6\textwidth}
	\rmfamily
	\vspace{-0.65\baselineskip}
	\centering
	\includegraphics[width=\linewidth]{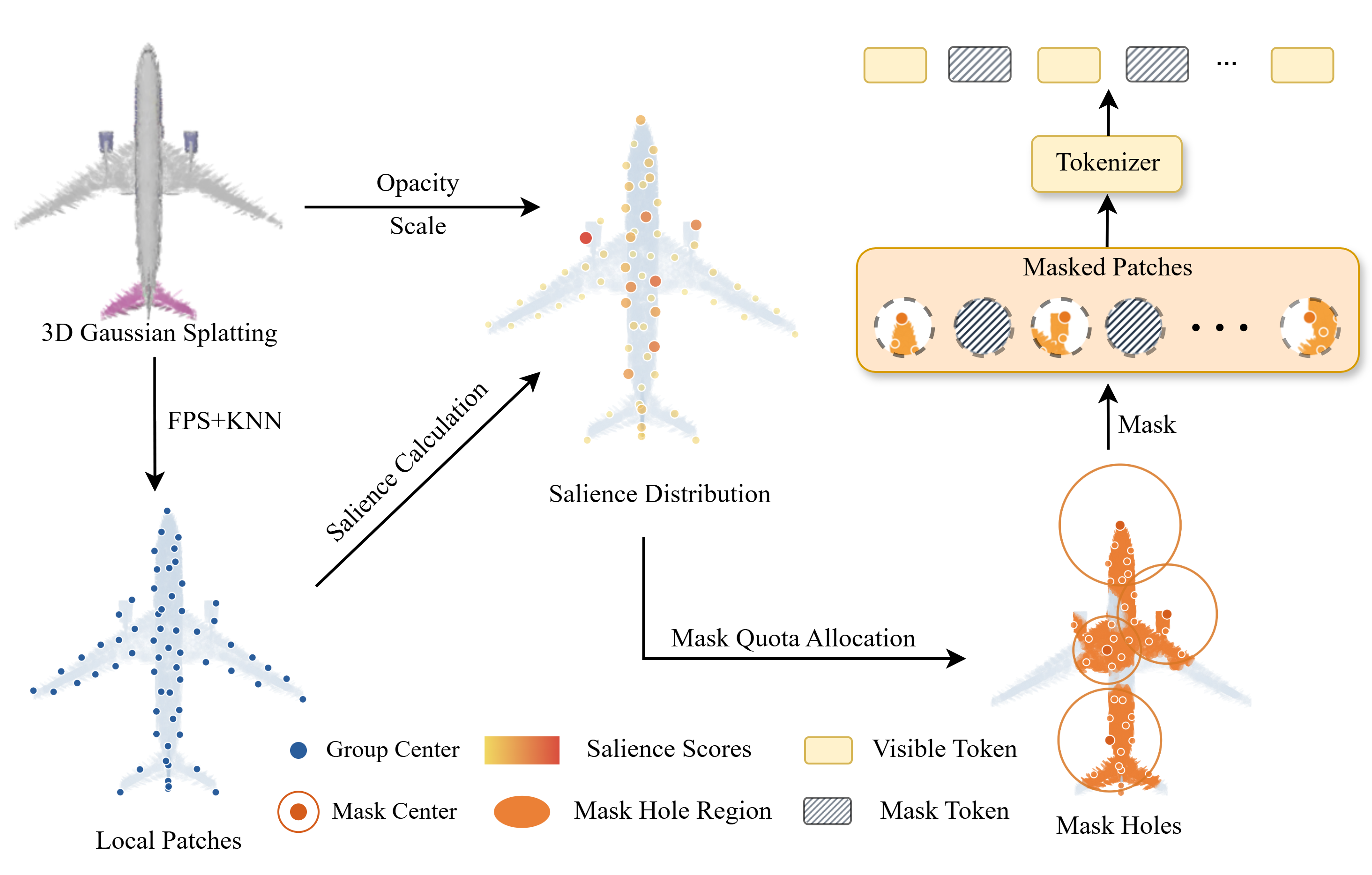}
	\caption{\textbf{Overview of the proposed GSHM strategy.} GaussFusion builds local patches from 3D Gaussian splats, computes salience scores for group centers, and allocates mask quotas by the salience distribution. Selected centers form multi-scale hole regions, where masked patches are replaced by mask tokens and the remaining patches serve as visible tokens for reconstruction.}
	\label{fig:mask}
	\vspace{-0.8\baselineskip}
\end{wrapfigure}

In masked pre-training for 3D Gaussian representations, the masking strategy affects not only reconstruction learning but also the use of multimodal supervision. Random masking usually selects local groups independently and produces scattered missing regions, which is not well suited to the non-uniform distribution of Gaussian primitives. In 3DGS, different primitives contribute unequally to geometry and appearance. Primitives with high opacity tend to have a stronger influence on rendered appearance, while primitives with large scales often cover broader local structures. Randomly removing isolated groups may therefore mask many low-information primitives or disrupt visually important regions, making the reconstruction target less useful for image- and text-guided semantic learning. 

To address this issue, we propose Gaussian Salience-guided Multi-scale Hole Masking (GSHM), as shown in Fig.~\ref{fig:mask}. GSHM constructs spatially continuous masked regions by jointly considering Gaussian salience and spatial coverage, thereby encouraging the encoder to recover semantically meaningful local structures from surrounding context.

GSHM first estimates the salience of each local Gaussian group from the attributes of its primitives. For the local group $\mathcal{N}_j$ with $K$ Gaussian primitives, opacity reflects its contribution to rendered appearance, while the scale parameters provide a proxy for the spatial coverage of the group. The salience score is computed as
\begin{equation}
	a_j
	=
	\left(
	\frac{1}{K}
	\sum_{k=1}^{K}
	\alpha_{j,k}
	\right)
	\cdot
	\left(
	\frac{1}{K}
	\sum_{k=1}^{K}
	s_{j,k}^{x}s_{j,k}^{y}s_{j,k}^{z}
	\right)^{\frac{1}{3}}.
\end{equation}

The first term measures the average visibility of the group, and the second term measures its average spatial extent using the geometric mean of the Gaussian scales. The salience scores are normalized within each sample and then mixed with a uniform distribution. This mixture prevents the masking process from collapsing to only a few highly salient regions, while still assigning higher sampling priority to groups that are more informative for appearance and structure.

After obtaining the salience-guided sampling weights $w_j$, GSHM selects multiple hole centers by salience-weighted farthest point sampling. The first center is sampled according to $w_j$. Given the previously selected hole centers $\mathcal{Q}_{h-1}$, the next center is selected by
\begin{equation}
	q_h
	=
	\arg\max_{j}
	\;
	w_j
	\cdot
	\min_{q \in \mathcal{Q}_{h-1}}
	\left\|
	\mathbf{p}_j-\mathbf{p}_{q}
	\right\|_2,
	\quad h=2,\ldots,H,
\end{equation}
where $\mathbf{p}_j$ denotes the spatial position of the $j$-th Gaussian group and $H$ is the number of hole centers. This criterion favors groups that are both salient and spatially distant from existing centers, producing diverse masked regions instead of concentrating all masks in a single local area.

Given the selected hole centers, each Gaussian group is assigned to its nearest center, forming a spatial partition over local Gaussian groups. The partition associated with the $h$-th hole center is defined as
\begin{equation}
	\mathcal{A}_h
	=
	\left\{
	j
	\;\middle|\;
	h
	=
	\arg\min_{r \in \{1,\ldots,H\}}
	\left\|
	\mathbf{p}_j-\mathbf{p}_{q_r}
	\right\|_2
	\right\}.
\end{equation}
GSHM then allocates different mask quotas to different hole centers. Instead of using identical hole sizes, a random scale perturbation is applied to the initial quota of each hole:
\begin{equation}
	\tilde{b}_h
	=
	\left\lfloor
	\frac{\rho G}{H}
	\cdot
	\eta_h
	\right\rceil,
	\quad
	\eta_h \sim \mathcal{U}(1-\gamma,1+\gamma),
\end{equation}
where $\rho$ denotes the mask ratio, $G$ is the number of local Gaussian groups, and $\gamma$ controls the degree of scale variation. The perturbed quotas are further adjusted so that the total number of masked groups matches the predefined mask ratio. This design produces multi-scale holes with different spatial extents and makes the reconstruction task closer to realistic incomplete observations.

Within each partition $\mathcal{A}_h$, GSHM selects the groups closest to the corresponding hole center as masked targets:
\begin{equation}
	\mathcal{M}_h
	=
	\operatorname{TopK}_{j\in\mathcal{A}_h}
	\left(
	-
	\left\|
	\mathbf{p}_j-\mathbf{p}_{q_h}
	\right\|_2,
	b_h
	\right),
\end{equation}
\begin{equation}
	\mathcal{M}
	=
	\bigcup_{h=1}^{H}
	\mathcal{M}_h,
\end{equation}
where $b_h$ is the adjusted mask quota of the $h$-th hole. If the union does not exactly satisfy the required number of masked groups because of empty partitions or quota perturbation, GSHM performs a final correction according to the group salience scores. The resulting mask therefore satisfies the predefined mask ratio while preserving spatial continuity and salience awareness. Compared with independent random masking, GSHM creates structured missing regions that encourage the model to infer semantically meaningful and spatially coherent Gaussian structures from visible context.

\subsection{Cross-Modal Semantic Alignment}\label{sec:3.2}

To improve the semantic discriminability of Gaussian representations, GaussFusion introduces image and text modalities as external supervision during pre-training. The image modality provides appearance-level cues from rendered views, while the text modality provides category-level semantic cues from textual descriptions. These modalities are used only as supervision targets. During downstream fine-tuning and inference, the model takes only 3D inputs and does not require additional image or text data.

Specifically, given a 3D Gaussian input $\mathcal{G}_i$, GaussFusion first divides it into local Gaussian groups and encodes the visible groups as a Gaussian token sequence $\mathbf{x}_i^{g}$. To inject multimodal supervision into the Gaussian encoder, two learnable tokens are prepended to the token sequence: an image alignment token $\mathbf{x}_i^{img}$ and a text alignment token $\mathbf{x}_i^{txt}$. These two tokens do not contain raw image or text content. Instead, they serve as learnable query tokens that collect 3D contextual information from visible Gaussian tokens through self-attention and are later constrained by external modality features. The input sequence of the Transformer encoder is defined as
\begin{equation}
	\mathbf{X}_i^{(0)}
	=
	[
	\mathbf{x}_i^{img};
	\mathbf{x}_i^{txt};
	\mathbf{x}_i^{g}
	]
	+
	\mathbf{P}_i,
\end{equation}
where $\mathbf{P}_i$ denotes the positional embedding. The encoded sequence is obtained by applying $L$ Transformer layers:
\begin{equation}
	\mathbf{X}_i^{(\ell)}
	=
	\operatorname{Trm}_{\ell}
	\left(
	\mathbf{X}_i^{(\ell-1)}
	\right),
	\quad
	\ell=1,\ldots,L.
\end{equation}
The final output sequence is denoted as
\begin{equation}
	\mathbf{U}_i
	=
	[
	\mathbf{u}_i^{img};
	\mathbf{u}_i^{txt};
	\mathbf{u}_i^{g}
	],
\end{equation}
where $\mathbf{u}_i^{img}$ and $\mathbf{u}_i^{txt}$ are the output features corresponding to the image and text alignment tokens, respectively.

The role of the alignment tokens can be interpreted from the self-attention operation. Let $m\in\{img,txt\}$ denote one of the two alignment tokens, and let $\mathcal{V}_i$ denote the visible Gaussian token indices of the $i$-th sample. In one attention head of the $\ell$-th Transformer layer, the contribution collected by the modality token from visible Gaussian tokens can be written as
\begin{equation}
	\mathbf{c}_{i,m}^{(\ell)}
	=
	\sum_{j\in\mathcal{V}_i}
	\omega_{i,mj}^{(\ell)}
	\mathbf{x}_{i,j}^{(\ell-1)}
	W_V^{(\ell)},
\end{equation}
where the attention weight is computed as
\begin{equation}
	\omega_{i,mj}^{(\ell)}
	=
	\frac{
		\exp
		\left(
		\frac{
			\left(
			\mathbf{x}_{i,m}^{(\ell-1)}W_Q^{(\ell)}
			\right)
			\left(
			\mathbf{x}_{i,j}^{(\ell-1)}W_K^{(\ell)}
			\right)^{\top}
		}{
			\sqrt{d}
		}
		\right)
	}{
		\sum_{r\in\Omega_i}
		\exp
		\left(
		\frac{
			\left(
			\mathbf{x}_{i,m}^{(\ell-1)}W_Q^{(\ell)}
			\right)
			\left(
			\mathbf{x}_{i,r}^{(\ell-1)}W_K^{(\ell)}
			\right)^{\top}
		}{
			\sqrt{d}
		}
		\right)
	},
\end{equation}
with $\Omega_i=\{img,txt\}\cup\mathcal{V}_i$. This formulation shows that the image and text alignment tokens aggregate information from visible Gaussian tokens through the same self-attention operation, rather than receiving image or text content directly inside the Gaussian encoder.

Each Gaussian sample $\mathcal{G}_i$ is paired with a rendered image $I_i$ and a text description $T_i$. The image is encoded by a frozen ViT image encoder\cite{dosovitskiy2021vit}, and the text description is encoded by a frozen CLIP text encoder\cite{radford2021clip}. Their output features are used as fixed semantic targets:
\begin{equation}
	\mathbf{z}_i^{img}
	=
	\operatorname{stopgrad}
	\left(
	F_{img}(I_i)
	\right),
\end{equation}
\begin{equation}
	\mathbf{z}_i^{txt}
	=
	\operatorname{stopgrad}
	\left(
	F_{txt}(T_i)
	\right),
\end{equation}
where $F_{img}(\cdot)$ and $F_{txt}(\cdot)$ denote the frozen image and text encoders, respectively. Since the image and text tokens produced by the Gaussian encoder are not directly in the same feature spaces as the external modality features, two linear projection layers $\phi_{img}(\cdot)$ and $\phi_{txt}(\cdot)$ are used for alignment:
\begin{equation}
	\hat{\mathbf{z}}_i^{img}
	=
	\phi_{img}
	\left(
	\mathbf{u}_i^{img}
	\right),
	\quad
	\hat{\mathbf{z}}_i^{txt}
	=
	\phi_{txt}
	\left(
	\mathbf{u}_i^{txt}
	\right).
\end{equation}

The projected image and text tokens are encouraged to match their corresponding external modality features through feature-level alignment. In this way, the image alignment token receives appearance supervision from rendered views, while the text alignment token receives semantic supervision from category descriptions. After pre-training, the image and text encoders are discarded, and the learned Gaussian encoder can be transferred to downstream 3D tasks without additional multimodal inputs.

\subsection{Reconstruction Target}

The reconstruction target of GaussFusion consists of a Gaussian masked reconstruction objective and a cross-modal semantic alignment objective.

Gaussian reconstruction preserves the core mechanism of Gaussian-MAE\cite{ma2025shapesplat}. The reconstruction objective contains multiple attribute terms, including position, opacity, scale, rotation, and spherical harmonics color. The reconstruction of spatial coordinates is constrained by Chamfer Distance, while the remaining attributes are constrained by the $\ell_1$ loss:
\begin{equation}
	\mathcal{L}_{rec}
	=
	\mathcal{L}_{xyz}
	+
	\mathcal{L}_{\alpha}
	+
	\mathcal{L}_{s}
	+
	\mathcal{L}_{q}
	+
	\mathcal{L}_{SH},
\end{equation}
where $\mathcal{L}_{xyz}$ denotes the coordinate reconstruction loss, while $\mathcal{L}_{\alpha}$, $\mathcal{L}_{s}$, $\mathcal{L}_{q}$, and $\mathcal{L}_{SH}$ denote the reconstruction losses for opacity, scale, rotation, and color attributes, respectively. The complete Gaussian reconstruction loss can be written as
\begin{equation}
	\mathcal{L}_{rec}
	=
	\mathcal{L}_{xyz}
	+
	\mathcal{L}_{attr}.
\end{equation}
This reconstruction objective forces the decoder to recover not only the spatial layout of masked regions, but also the appearance-related attributes carried by Gaussian primitives.

The cross-modal semantic alignment objective constrains the image-aligned and text-aligned representations produced by the Gaussian branch. Frozen image and text encoders are used, and their output features serve as semantic targets. The Gaussian branch predicts the semantic features of the corresponding modalities through the image token and the text token, and Smooth L1 loss is used to reduce the discrepancy between the predicted features and the target features. The image alignment loss and the text alignment loss are defined as
\begin{equation}
	\mathcal{L}_{img}
	=
	\operatorname{SmoothL1}
	\left(
	\mathbf{z}_i^{img},
	\hat{\mathbf{z}}_i^{img}
	\right),
\end{equation}
\begin{equation}
	\mathcal{L}_{txt}
	=
	\operatorname{SmoothL1}
	\left(
	\mathbf{z}_i^{txt},
	\hat{\mathbf{z}}_i^{txt}
	\right).
\end{equation}

Finally, the overall optimization objective of GaussFusion is defined as
\begin{equation}
	\mathcal{L}
	=
	\mathcal{L}_{rec}
	+
	\lambda_{\mathrm{img}}\mathcal{L}_{img}
	+
	\lambda_{\mathrm{text}}\mathcal{L}_{txt}.
\end{equation}

\section{Experiments}\label{sec:experiments}

We evaluate GaussFusion by testing whether the pre-trained Gaussian representations can transfer to different 3D understanding tasks. The evaluation covers real-world scanned object classification, clean CAD object classification, fine-grained part segmentation, and few-shot classification. ScanObjectNN\cite{uy2019scanobjectnn} tests robustness under background clutter and perturbations, ModelNet10/40\cite{wu2015modelnet} measure standard object-level recognition, ShapeNetPart\cite{yi2016shapenetpart} examines local part understanding, and few-shot classification reflects transferability under limited annotations.

The experiments also isolate the main design choices in GaussFusion. The ablation study examines how image and text supervision, GSHM masking, encoder selection, and multimodal loss weights affect downstream performance.

\begin{table}[t]
	\rmfamily
	\centering
	\caption{\textbf{Object classification on ScanObjectNN.} Overall accuracy (\%) is reported. $^{\dagger}$ denotes reproduced results under our implementation.}
	\label{tab:scanobjectnn_cls}
	\setlength{\tabcolsep}{3pt}
	\renewcommand{\arraystretch}{0.7}
	\begin{tabularx}{\linewidth}{@{}l*{3}{>{\centering\arraybackslash}X}@{}}
		\toprule
		Method & OBJ-BG & OBJ-ONLY & PB-T50-RS \\
		\midrule
		\multicolumn{4}{c}{\textit{with Point Cloud Pre-training Representation}} \\
		\midrule
		PointNet\cite{qi2017pointnet} & 73.30 & 79.20 & 68.00 \\
		SpiderCNN\cite{xu2018spidercnn} & 77.10 & 79.50 & 73.70 \\
		PointNet++\cite{qi2017pointnetplusplus} & 82.30 & 84.30 & 77.90 \\
		DGCNN\cite{wang2019dgcnn} & 82.80 & 86.20 & 78.10 \\
		PointCNN\cite{li2018pointcnn} & 86.10 & 85.50 & 78.50 \\
		Transformer\cite{yu2022pointbert} & 79.86 & 80.55 & 77.24 \\
		Transformer + OcCo\cite{yu2022pointbert} & 84.85 & 85.54 & 78.79 \\
		Point-MAE\cite{pang2022pointmae} & 90.02 & 88.29 & 85.18 \\
		Point-M2AE\cite{zhang2022pointm2ae} & 91.22 & 88.81 & 86.43 \\
		Mamba3D\cite{han2024mamba3d} & 92.94 & 92.08 & 91.81 \\
		\midrule
		\multicolumn{4}{c}{\textit{with 3D Gaussian Pre-training Representation}} \\
		\midrule
		Gaussian-MAE$^{\dagger}$\cite{ma2025shapesplat} & 84.30 & 86.23 & 78.87 \\
		\rowcolor{lightgreen}
		GaussFusion & \textbf{88.98(\textcolor{mygreen}{+4.68})} & \textbf{88.47(\textcolor{mygreen}{+2.24})} & \textbf{82.72(\textcolor{mygreen}{+3.85})} \\
		\bottomrule
	\end{tabularx}
\end{table}

\begin{figure}
	\centering
	\includegraphics[width=0.95\linewidth,height=0.6\linewidth]{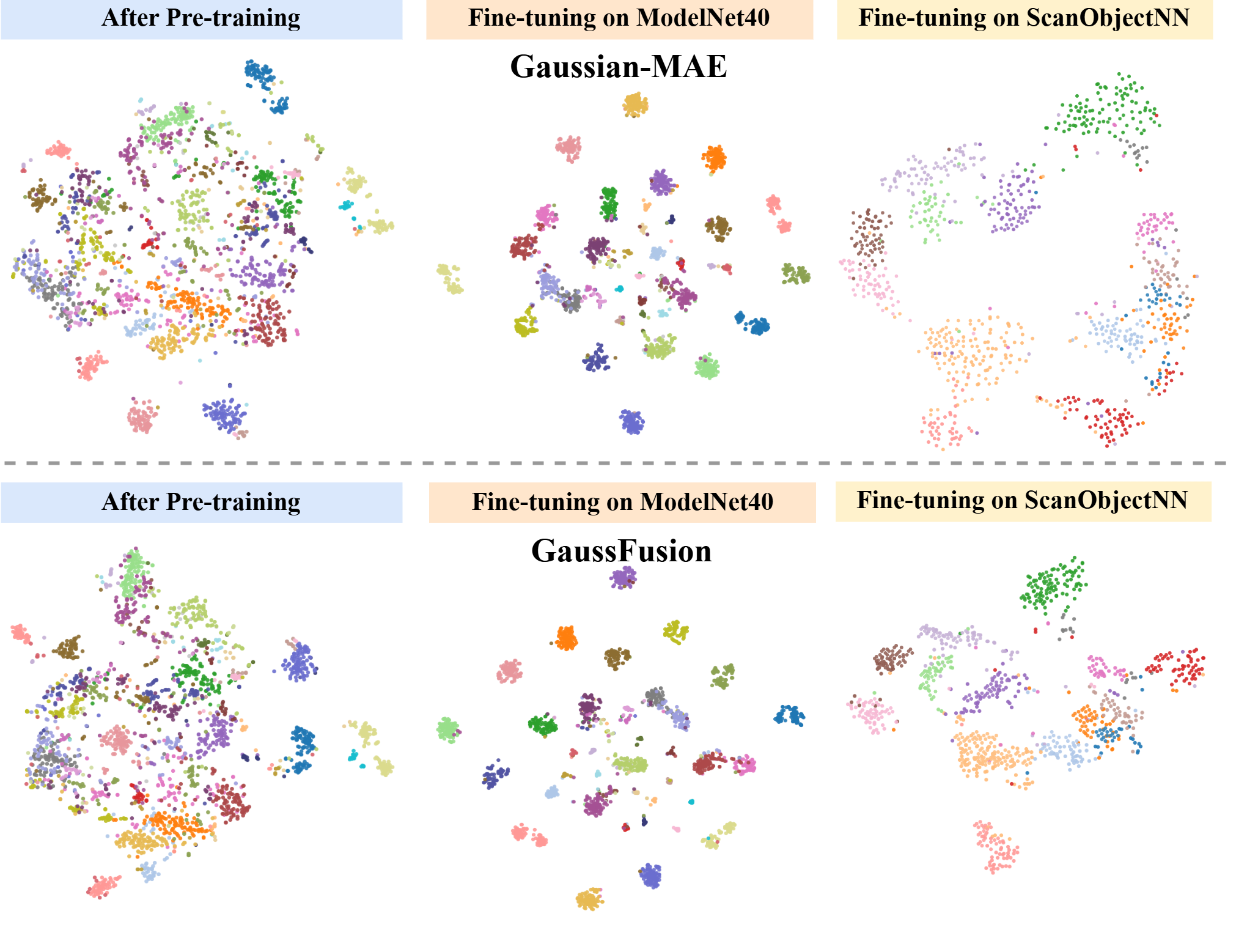}
	\caption{\textbf{Visualization of feature distributions.} The t-SNE plots compare the feature embeddings of Gaussian-MAE and GaussFusion after pre-training and after fine-tuning on ModelNet40 and ScanObjectNN. Different colors represent different object categories.}\label{fig:tsne}
\end{figure}
\subsection{Pre-training Setup}
\begin{table}[t]
	\rmfamily
	\centering
	\caption{\textbf{Object classification on clean ModelNet datasets.} Input indicates the representation used by each method. Overall accuracy (\%) is reported. $^{\dagger}$ denotes reproduced results under our implementation.}
	\label{tab:modelnet_cls}
	\setlength{\tabcolsep}{2pt}
	\renewcommand{\arraystretch}{0.7}
	\begin{tabularx}{\linewidth}{@{}l*{3}{>{\centering\arraybackslash}X}@{}}
		\toprule
		Method & Input & ModelNet10 & ModelNet40 \\
		\midrule
		\multicolumn{4}{c}{\textit{with Point Cloud Pre-training Representation}} \\
		\midrule
		PointNet\cite{qi2017pointnet} & xyz & 94.40 & 89.20 \\
		PointNet++\cite{qi2017pointnetplusplus} & xyz+normal & 94.10 & 91.90 \\
		DGCNN\cite{wang2019dgcnn} & xyz & 95.00 & 92.20 \\
		PointCNN\cite{li2018pointcnn} & xyz & -- & 92.50 \\
		PCNN\cite{atzmon2018pcnn} & xyz & 94.90 & 92.30 \\
		A-CNN\cite{komarichev2019acnn} & xyz+normal & 95.50 & 92.60 \\
		PointASNL\cite{yan2020pointasnl} & xyz & 95.70 & 92.90 \\
		P2SResLNet\cite{wu2024p2sreslnet} & xyz & -- & 90.60 \\
		Point Cloud Mamba\cite{zhang2024pointcloudmamba} & xyz & -- & 93.40 \\
		Mamba3D\cite{han2024mamba3d} & xyz & -- & 93.40 \\
		\midrule
		\multicolumn{4}{c}{\textit{with 3D Gaussian Pre-training Representation}} \\
		\midrule
		Gaussian-MAE$^{\dagger}$\cite{ma2025shapesplat} & Gaussian & 95.37 & 92.46 \\
		\rowcolor{lightgreen}
		GaussFusion & Gaussian & \textbf{95.82(\textcolor{mygreen}{+0.45})} & \textbf{93.07(\textcolor{mygreen}{+0.61})} \\
		\bottomrule
	\end{tabularx}
\end{table}

GaussFusion is pre-trained on the ShapeSplat dataset in a self-supervised manner. ShapeSplat\cite{ma2025shapesplat} is a large-scale 3DGS dataset that contains approximately 52,000 high-quality 3D Gaussian Splatting models. Each object is represented in the form of Gaussian parameters, including Gaussian attributes such as center position, opacity, scale, rotation quaternion, and spherical harmonics color coefficients.

To enable multimodal learning, each object in ShapeSplat is also equipped with image and text semantic information. Multi-view images are rendered from 16 predefined viewpoints and contain object silhouettes, texture cues, occlusion relations, and appearance consistency under viewpoint changes. Text descriptions are composed of category names and templated attribute phrases, providing explicit category and attribute semantics for Gaussian representations.

In GaussFusion, 1024 Gaussian primitives are randomly sampled from each object during training. The input Gaussians are divided into 128 local groups, each containing 32 primitives. Grouping is performed based on xyz coordinates, while the local encoder uses complete Gaussian attributes for feature extraction. The Transformer encoder has a feature dimension of 384 and consists of 12 layers with 6 attention heads. The decoder consists of 4 Transformer blocks with 6 attention heads.

The mask ratio is set to 0.6 during pre-training. GSHM generates 4 spatial masking regions for each sample, and the salience mixing coefficient and multi-scale jitter coefficient are set to 0.7 and 0.35, respectively. The model reconstructs the masked xyz, opacity, scale, rotation, and SH attributes. Chamfer Distance is used for xyz reconstruction, while the $\ell_1$ loss is used for the remaining attributes. Cross-modal supervision uses a frozen ViT-B/32 image encoder\cite{dosovitskiy2021vit} and a frozen CLIP text encoder\cite{radford2021clip}. Both image and text alignment losses adopt Smooth L1 loss with a weight of 1.0.

All pre-training experiments use the AdamW optimizer with an initial learning rate of 0.001 and a weight decay of 0.05. The model is trained for 300 epochs with a cosine learning rate schedule and 10 warm-up epochs. The total batch size is 128.

\subsection{Downstream Tasks}
For downstream evaluation, only the pre-trained Gaussian encoder is transferred. The image and text encoders are used during pre-training only and are removed during fine-tuning and inference. This protocol ensures that GaussFusion does not introduce additional multimodal inputs or inference cost in downstream tasks.
\subsubsection{Classification Experiments}

\textbf{Object Classification on Real-World Datasets.} For real-world object classification, ScanObjectNN\cite{uy2019scanobjectnn} is used to evaluate the classification ability of different models on real scanned data. Unlike clean CAD models, this dataset contains background clutter, occlusion, and pose perturbations, making it more suitable for testing the robustness of pre-trained representations in real scenarios. It also helps examine whether a method overfits standard data characteristics or learns more general semantic representations. Table~\ref{tab:scanobjectnn_cls} reports the classification results on the OBJ-BG, OBJ-ONLY, and PB-T50-RS splits. Fig.~\ref{fig:tsne} provides a t-SNE visualization of the feature distributions after pre-training and after fine-tuning on ModelNet40 and ScanObjectNN, illustrating the evolution from mixed embeddings to more discriminative category clusters.

\begin{figure*}
	\rmfamily
	\centering
	\includegraphics[width=\linewidth]{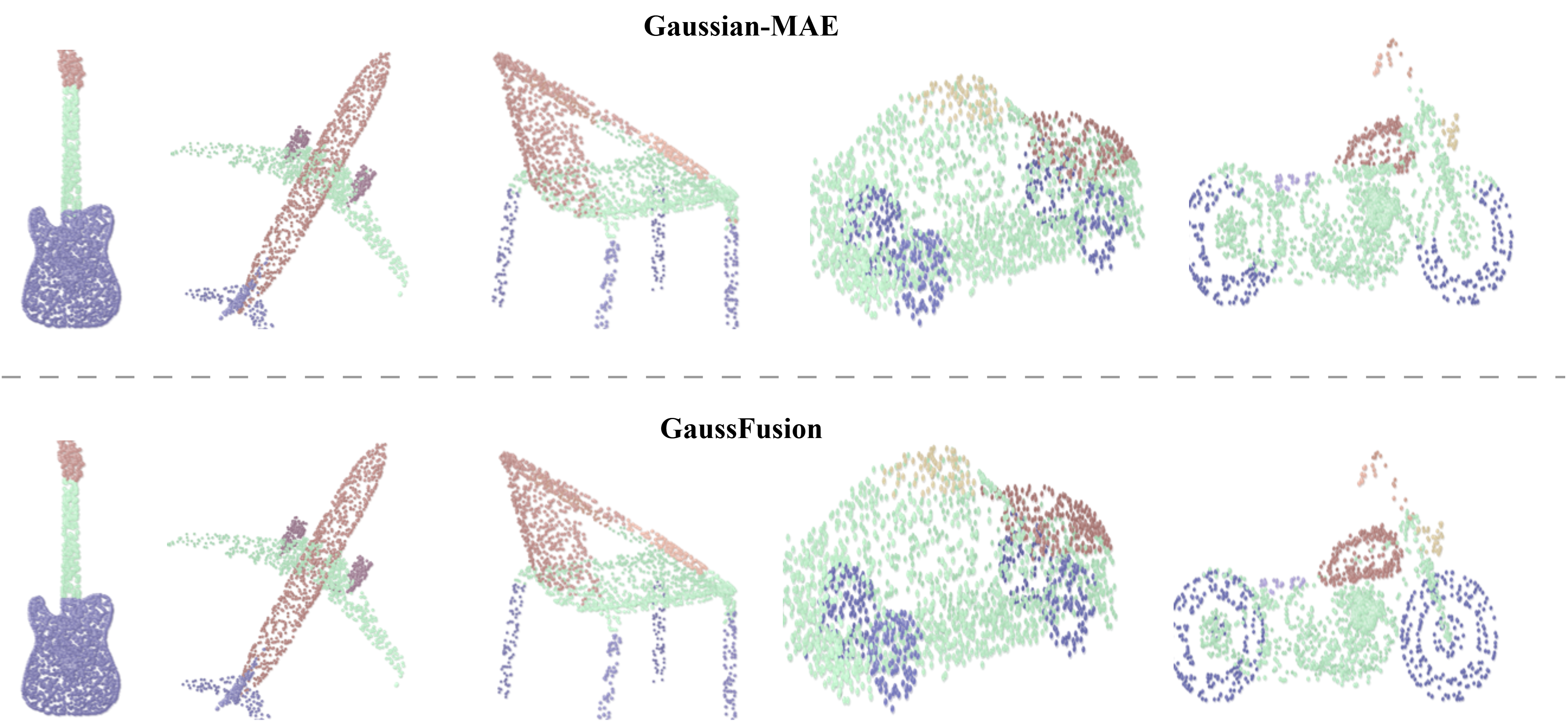}
	\caption{\textbf{Qualitative comparison of part segmentation results on ShapeNetPart.} Different colors represent predicted semantic parts. Compared with Gaussian-MAE, GaussFusion produces more spatially coherent predictions and clearer part boundaries, particularly for thin structures and geometrically complex regions.}
	\label{fig:part_seg}
\end{figure*}

Compared with point cloud pre-training methods such as Point-MAE\cite{pang2022pointmae}, GaussFusion still shows a performance gap. This is mainly related to the difference in input representation and pre-training data form. Point-MAE is pre-trained and transferred directly in the point cloud space, whereas GaussFusion uses 3D Gaussian attributes as the pre-training target and then transfers the learned representation to point cloud classification. This setting introduces the difficulty of cross-representation transfer. Therefore, the comparison with Gaussian-MAE\cite{ma2025shapesplat} under the same Gaussian representation setting is more central to this work.

Under the Gaussian representation setting, GaussFusion outperforms Gaussian-MAE\cite{ma2025shapesplat} on the splits with background clutter and strong perturbations. In particular, it achieves 82.72\% on PB-T50-RS, the most challenging setting, improving over Gaussian-MAE by 3.85\%. This result indicates that multimodal supervision and the GSHM masking strategy help the model learn more transferable Gaussian representations.

\textbf{Object Classification on Clean Object Datasets.} ModelNet10 and ModelNet40\cite{wu2015modelnet} are used to evaluate classification performance on clean object data. These datasets mainly consist of regular CAD models with limited background noise and scanning incompleteness. They therefore serve as a complement to ScanObjectNN and are used to examine whether GaussFusion remains effective on objects with clear structures. Table~\ref{tab:modelnet_cls} presents comparisons with classical point cloud classification methods and Gaussian-MAE.

On both ModelNet datasets, GaussFusion outperforms Gaussian-MAE\cite{ma2025shapesplat}. This result shows that the proposed pre-training strategy not only improves robustness in real-world scenarios, but also enhances category discrimination on clean object data. Since some point cloud methods in the table use different input forms and training protocols, they are mainly included as reference baselines. A more direct comparison is between Gaussian-MAE and GaussFusion, as both adopt the same Gaussian input setting and thus more clearly reflect the effects of multimodal guidance and the GSHM masking strategy.

\subsubsection{Segmentation Experiments}

\begin{table}
	\rmfamily
	\centering
	\setlength{\tabcolsep}{0.6pt}
	\renewcommand{\arraystretch}{0.6}
	\caption{\textbf{Part segmentation results on the ShapeNetPart dataset.} We report the mean IoU across all part categories mIoU$_C$ (\%) and the mean IoU across all instances mIoU$_I$ (\%), as well as the IoU (\%) for each category. $^{\dagger}$ denotes reproduced results under our implementation.}
	\label{tab:part_segmentation}
	\begin{tabular*}{\textwidth}{@{\extracolsep{\fill}} lcc cccccccc @{}}
		\toprule
		Method & mIoU$_C$ & mIoU$_I$ & aero & bag & cap & car & chair & e-phone & guitar & knife \\
		& & & lamp & laptop & motor & mug & pistol & rocket & s-board & table \\
		\midrule
		PointNet\cite{qi2017pointnet} & 80.39 & 83.7 & 83.4 & 78.7 & 82.5 & 74.9 & 89.6 & 73.0 & 91.5 & 85.9 \\
		& & & 80.8 & 95.3 & 65.2 & 93.0 & 81.2 & 57.9 & 72.8 & 80.6 \\
		PointNet++\cite{qi2017pointnetplusplus} & 81.85 & 85.1 & 82.4 & 79.0 & 87.7 & 77.3 & 90.8 & 71.8 & 91.0 & 85.9 \\
		& & & 83.7 & 95.3 & 71.6 & 94.1 & 81.3 & 58.7 & 76.4 & 82.6 \\
		DGCNN\cite{wang2019dgcnn} & 82.33 & 85.2 & 84.0 & 83.4 & 86.7 & 77.8 & 90.6 & 74.7 & 91.2 & 87.5 \\
		& & & 82.8 & 95.7 & 66.3 & 94.9 & 81.1 & 63.5 & 74.5 & 82.6 \\
		Transformer\cite{yu2022pointbert} & 83.42 & 85.1 & 82.9 & 85.4 & 87.7 & 78.8 & 90.5 & 80.8 & 91.1 & 87.7 \\
		& & & 85.3 & 95.6 & 73.9 & 94.9 & 83.5 & 61.2 & 74.9 & 80.6 \\
		Point-BERT\cite{yu2022pointbert} & 84.11 & 85.6 & 84.3 & 84.8 & 88.0 & 79.8 & 91.0 & 81.7 & 91.6 & 87.9 \\
		& & & 85.2 & 95.6 & 75.6 & 94.7 & 84.3 & 63.4 & 76.3 & 81.5 \\
		Point-MAE\cite{pang2022pointmae} & 84.19 & 86.1 & 84.3 & 85.0 & 88.3 & 80.5 & 91.3 & 78.5 & 92.1 & 87.4 \\
		& & & 86.1 & 96.1 & 75.2 & 94.6 & 84.7 & 63.5 & 77.1 & 82.4 \\
		Gaussian-MAE$^{\dagger}$\cite{ma2025shapesplat} & 83.65 & 85.8 & 84.8 & 84.8 & 90.2 & 80.2 & 90.8 & 73.1 & 91.8 & 87.6 \\
		& & & 84.4 & 95.9 & 76.6 & 95.9 & 85.3 & 61.5 & 73.9 & 81.9 \\
		\rowcolor{lightgreen}
		GaussFusion & \textbf{84.43} & \textbf{86.3} & 85.2 & 84.2 & 87.7 & 80.4 & 91.4 & 72.6 & 92.2 & 87.5 \\
		\rowcolor{lightgreen}
		& & & 84.4 & 95.9 & 78.8 & 95.5 & 84.8 & 65.8 & 75.6 & 82.2 \\
		\bottomrule
	\end{tabular*}
\end{table}
ShapeNetPart\cite{yi2016shapenetpart} is used to evaluate fine-grained local structure understanding. Unlike object classification, part segmentation requires the model not only to recognize the object category, but also to predict the corresponding semantic part label for each point. Therefore, this task further examines whether the pre-trained Gaussian representations preserve local geometry and part-level semantic information.

The experiments use the same dataset split and part annotations as those used by point cloud methods. Since the input is a 3D Gaussian representation while the supervision signals in ShapeNetPart are defined on point cloud positions, Gaussian data are aligned with the original point cloud annotations according to the mapping files, and segmentation results are computed at the annotated point cloud positions. This protocol avoids unfair comparisons caused by different supervision locations. Table~\ref{tab:part_segmentation} reports the segmentation results of different methods on ShapeNetPart. Fig.~\ref{fig:part_seg} provides qualitative visualization results of our part segmentation predictions.

The results show that GaussFusion outperforms Gaussian-MAE in both mIoU$_C$ and mIoU$_I$, indicating improved transferability of the Gaussian encoder to part-level prediction. Across categories, GaussFusion achieves better results on categories containing thin parts or local structural variations, suggesting that the GSHM masking strategy encourages the model to focus on more discriminative local regions. Meanwhile, GaussFusion does not surpass Gaussian-MAE on several simple categories. A possible reason is that, for categories with simple shapes, limited training samples, or ambiguous part boundaries, image- and text-level supervision tends to emphasize global semantics and provides limited help for fine-grained part boundaries.

\subsubsection{Few-shot Classification}
\begin{table*}[t]
	\rmfamily
	\centering
	\setlength{\tabcolsep}{0.05pt}
	\renewcommand{\arraystretch}{0.7}
	\caption{\textbf{Few-shot object classification on ModelNet40.} We report the average accuracy (\%) and standard deviation (\%) of 10 independent experiments. $^{\dagger}$ denotes reproduced results under our implementation. Bold denotes the best result among Gaussian-representation methods.}
	\label{tab:fewshot}
	
	\begin{tabularx}{\textwidth}{
			l*{4}{>{\centering\arraybackslash}X}		}
		\toprule
		Method
		& 5-way, 10-shot
		& 5-way, 20-shot
		& 10-way, 10-shot
		& 10-way, 20-shot \\
		\midrule
		
		\multicolumn{5}{c}{\textit{with Point Cloud Pre-training Representation}} \\
		\midrule
		DGCNN-rand\cite{wang2021occo}
		& $31.6 \pm 2.8$ & $40.8 \pm 4.6$ & $19.9 \pm 2.1$ & $16.9 \pm 1.5$ \\
		+OcCo\cite{wang2021occo}
		& $90.6 \pm 2.8$ & $92.5 \pm 1.9$ & $82.9 \pm 1.3$ & $86.5 \pm 2.2$ \\
		Transformer\cite{yu2022pointbert}
		& $87.8 \pm 5.2$ & $93.3 \pm 4.3$ & $84.6 \pm 5.5$ & $89.4 \pm 6.3$ \\
		+OcCo\cite{yu2022pointbert}
		& $94.0 \pm 3.6$ & $95.9 \pm 2.3$ & $89.4 \pm 5.1$ & $92.4 \pm 4.6$ \\
		Point-BERT\cite{yu2022pointbert}
		& $94.6 \pm 3.1$ & $96.3 \pm 2.7$ & $91.0 \pm 5.4$ & $92.7 \pm 5.1$ \\
		Point-MAE\cite{pang2022pointmae}
		& $96.3 \pm 2.5$ & $97.8 \pm 1.8$ & $92.6 \pm 4.1$ & $95.0 \pm 3.0$ \\
		Mamba3D\cite{han2024mamba3d}
		& $96.4 \pm 2.2$ & $98.2 \pm 1.2$ & $92.4 \pm 4.1$ & $95.2 \pm 2.9$ \\
		\midrule
		
		\multicolumn{5}{c}{\textit{with 3D Gaussian Pre-training Representation}} \\
		\midrule
		Gaussian-MAE$^{\dagger}$\cite{ma2025shapesplat}
		& $94.0 \pm 3.5$ & $95.1 \pm 3.6$ & $87.3 \pm 6.1$ & $92.2 \pm 4.7$ \\
		
		\rowcolor{lightgreen}
		GaussFusion
		& {\small $\mathbf{96.1(\textcolor{mygreen}{+2.1}) \pm 2.4}$}
		& {\small $\mathbf{96.5(\textcolor{mygreen}{+1.4}) \pm 2.3}$}
		& {\small $\mathbf{88.9(\textcolor{mygreen}{+1.6}) \pm 5.9}$}
		& {\small $\mathbf{93.1(\textcolor{mygreen}{+0.9}) \pm 4.9}$} \\
		\bottomrule
	\end{tabularx}
\end{table*}

Few-shot classification experiments evaluate the transferability of pre-trained representations in low-annotation scenarios and directly assess whether the model learns transferable category semantics. The experiments are conducted on ModelNet40\cite{wu2015modelnet} under 5-way and 10-way settings, with 10-shot and 20-shot training samples, respectively. Each setting is independently run 10 times, and the mean accuracy and standard deviation are reported. Table~\ref{tab:fewshot} presents the few-shot classification results.

Since the point cloud methods in Table~\ref{tab:fewshot} are pre-trained and transferred directly in the point cloud space, their results are included mainly as reference baselines. The primary comparison is therefore conducted with Gaussian-MAE\cite{ma2025shapesplat}, which adopts the same 3D Gaussian representation and downstream evaluation protocol as GaussFusion. GaussFusion consistently outperforms Gaussian-MAE across all four few-shot settings, indicating that multimodal semantic supervision and GSHM improve the transferability of Gaussian representations under limited annotations. The improvement is more pronounced under the 5-way settings, which may indicate that when the number of categories is smaller, the semantic priors provided by image and text supervision can more effectively help the model form clear inter-class boundaries. Therefore, the few-shot experiments further demonstrate that the benefits of GaussFusion come not only from fine-tuning under fully annotated conditions, but also from transfer in low-annotation scenarios.

\subsection{Ablation Study}
\begin{table}[t]
	\rmfamily
	\centering
	\renewcommand{\arraystretch}{0.7}
	\begin{minipage}[c]{0.61\textwidth}
		\centering
		\refstepcounter{table}\label{tab:multimodal_input_ablation}
		\parbox{\linewidth}{
			\textbf{Table~\thetable}\\
			\textbf{Ablation study of GSHM and cross-modal supervision on ScanObjectNN OBJ-BG.}
			Overall accuracy (\%) is reported.
		}
		\vspace{2pt}
		
		\begin{tabular*}{\linewidth}{@{\extracolsep{\fill}}cccc@{}}
			\toprule
			\multirow{2}{*}{\makecell{GSHM\\Mask}} & \multicolumn{2}{c}{Cross-modal Inputs} & \multirow{2}{*}{ScanObjectNN} \\
			\cmidrule(lr){2-3}
			& Image Input & Text Input &  \\
			\midrule
			\xmark & \xmark & \xmark & 84.30 \\
			\xmark & \cmark & \xmark & 85.89 \\
			\xmark & \xmark & \cmark & 86.23 \\
			\xmark & \cmark & \cmark & 87.95 \\
			\midrule
			\cmark & \xmark & \xmark & 85.03 \\
			\cmark & \cmark & \xmark & 86.91 \\
			\cmark & \xmark & \cmark & 87.43 \\
			\cmark & \cmark & \cmark & \textbf{88.98} \\
			\bottomrule
		\end{tabular*}
	\end{minipage}
	\hfill
	\begin{minipage}[c]{0.35\textwidth}
		\centering
		\refstepcounter{table}\label{tab:encoder_ablation}
		\parbox{\linewidth}{
			\textbf{Table~\thetable}\\
			\textbf{Ablation study of different image encoders on ScanObjectNN.}
			The text encoder is fixed to CLIP-B. Overall accuracy (\%) is reported.
		}
		\vspace{2pt}
		
		\begin{tabular*}{\linewidth}{@{\extracolsep{\fill}}cc@{}}
			\toprule
			Image Encoder & Acc. (\%) \\
			\midrule
			BEiT-B\cite{bao2022beit} & 87.61 \\
			ResNet-50\cite{he2016resnet} & 87.95 \\
			CLIP-B\cite{radford2021clip} & 88.30 \\
			Swin-B\cite{liu2021swin} & 88.64 \\
			ViT-B\cite{dosovitskiy2021vit} & \textbf{88.98} \\
			\bottomrule
		\end{tabular*}
	\end{minipage}
\end{table}
\subsubsection{Effect of Multimodal Supervision}

The roles of image and text supervision are evaluated under both random masking and GSHM. As shown in Table~\ref{tab:multimodal_input_ablation}, introducing either modality consistently improves classification performance, while combining the two leads to further gains under both masking strategies. Text supervision brings a slightly larger improvement than image supervision. One possible reason is that the text input is derived from category descriptions, which directly provide object-level semantic constraints. In contrast, the image input is rendered from a single view or a limited number of views and is more susceptible to variations in viewpoint, occlusion, and appearance. The best result is achieved when image and text supervision are used together, indicating that the two modalities provide complementary guidance and jointly encourage the Gaussian encoder to learn more transferable representations.

\subsubsection{Effect of Masking Strategy}

\begin{figure}
	\rmfamily
	\centering
	\includegraphics[width=0.95\linewidth]{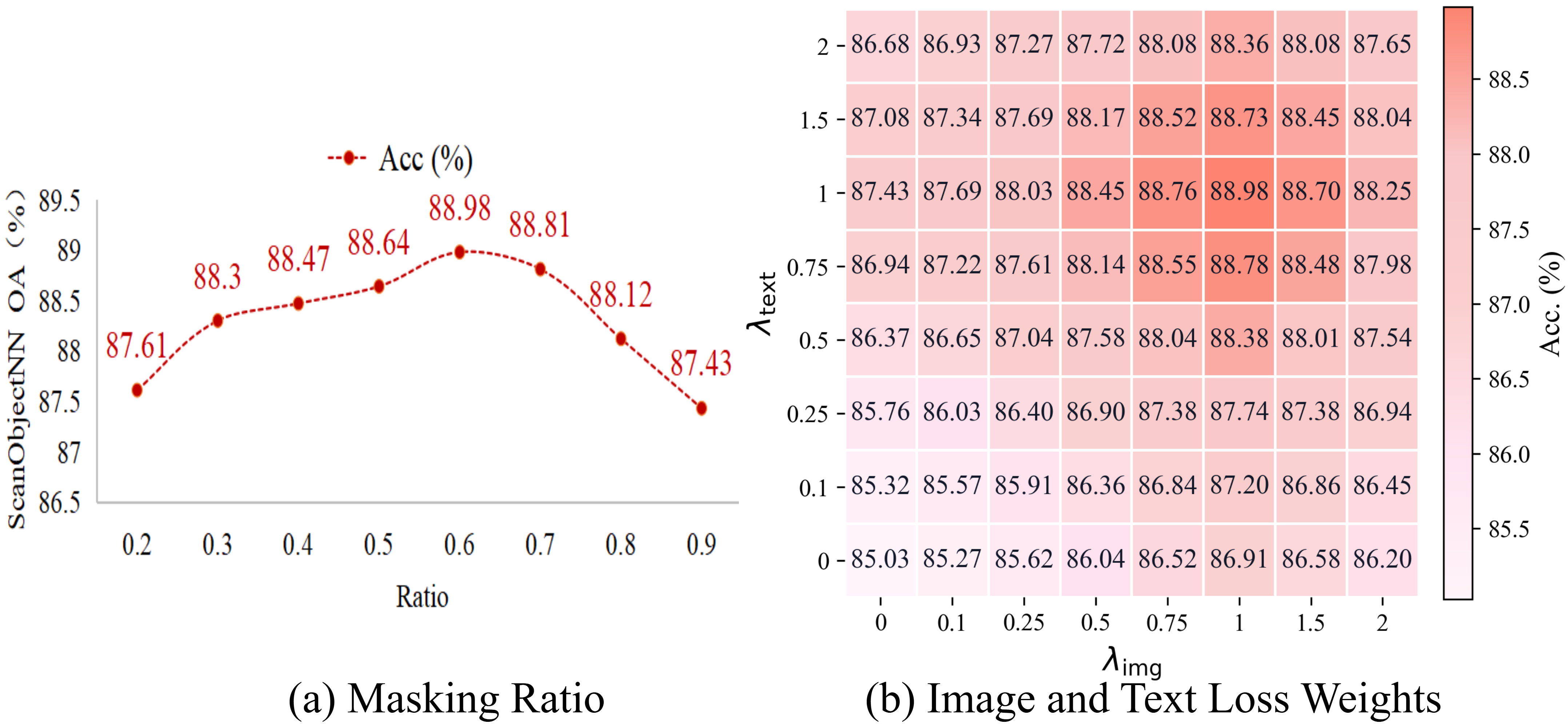}
	\caption{\textbf{Ablation study on masking ratio and multimodal loss weights.} 
		(a) shows the effect of GSHM masking ratios on ScanObjectNN OBJ-BG. 
		(b) shows the effect of image and text loss weights, with the best accuracy of 88.98\% at $\lambda_{\mathrm{img}}=\lambda_{\mathrm{text}}=1$.}\label{fig:lamda}
\end{figure}

The effectiveness of GSHM is evaluated by comparing it with random masking under the same cross-modal supervision settings. Table~\ref{tab:multimodal_input_ablation} shows that GSHM consistently improves classification performance across all image and text input combinations, indicating that spatially coherent masking provides more effective reconstruction targets for multimodal Gaussian pre-training. Fig.~\ref{fig:lamda}(a) further shows that performance gradually improves as the masking ratio increases from 0.2 to 0.6 and decreases when the ratio is further increased. This result suggests that a moderate masking ratio provides an appropriate balance between reconstruction difficulty and visible context. Therefore, the masking ratio is set to 0.6 in the main experiments.

\subsubsection{Effect of Different Encoders}

This section analyzes the influence of different image encoders on GaussFusion. The Gaussian reconstruction objective, text encoder\cite{radford2021clip}, and GSHM masking strategy are kept unchanged, while only the pre-trained encoder in the image branch is replaced. Table~\ref{tab:encoder_ablation} shows that all evaluated visual encoders provide effective supervision for Gaussian pre-training, with ViT achieving the best result.

Although the text branch uses the CLIP text encoder\cite{radford2021clip}, the image branch adopts ViT\cite{dosovitskiy2021vit} as the image encoder rather than the CLIP image encoder. It is observed that the image and text encoders in CLIP are jointly trained in the same image-text alignment space, so their supervision signals may have strong overlap. In contrast, the ViT image encoder provides more independent visual structural features. When combined with the CLIP text encoder, the image and text branches form a clearer complementary relationship.

\subsubsection{Hyperparameter Analysis}

The sensitivity to multimodal loss weights is studied by varying the image alignment weight $\lambda_{\mathrm{img}}$ and the text alignment weight $\lambda_{\mathrm{text}}$. The Gaussian reconstruction objective, GSHM masking strategy, and other training settings are kept unchanged. Fig.~\ref{fig:lamda} shows the performance landscape under different weight combinations.

The results show that when the two cross-modal loss weights are small, the model mainly relies on the Gaussian reconstruction objective and receives insufficient semantic supervision, leading to relatively low classification performance on ScanObjectNN\cite{uy2019scanobjectnn}. As $\lambda_{\mathrm{img}}$ and $\lambda_{\mathrm{text}}$ increase, the model performance gradually improves, indicating that image and text features provide effective semantic constraints for Gaussian representation learning. The best result is achieved when both weights are set to 1. Further increasing the weights leads to performance degradation, suggesting that overly strong cross-modal alignment shifts the training objective toward the image-text semantic space and weakens the focus on local Gaussian geometry and attribute reconstruction. Based on this analysis, both $\lambda_{\mathrm{img}}$ and $\lambda_{\mathrm{text}}$ are set to 1 in the main experiments.

\section{Conclusion}\label{sec:conclusion}
In this paper, we propose GaussFusion, a multimodal self-supervised pre-training framework for 3D Gaussian representations. By combining masked Gaussian reconstruction with image and text supervision, GaussFusion learns representations that preserve local Gaussian structure while absorbing high-level semantic information from external modalities. We further introduce GSHM, a salience-guided multi-scale masking strategy designed for the non-uniform distribution of Gaussian primitives. Experiments on classification, segmentation, and few-shot recognition show that GaussFusion improves the transferability of Gaussian representations across different downstream tasks. Current multimodal supervision is mainly derived from rendered images and category-level text, and future work will explore richer semantic annotations and larger scene-level Gaussian representations.
\section*{CRediT authorship contribution statement}
Zhixuan You: Conceptualization, Methodology, Writing - Original Draft. Jihua Zhu: Software, Investigation. Yiding Sun: Resources, Visualization. Zihao Guo: Investigation. Haozhe Cheng: Writing - Review and Editing. Dongxu Zhang: Writing - Review and Editing. Lin Chen: Writing - Review and Editing. Hainan Luo: Writing - Review and Editing.

\bibliographystyle{elsarticle-num}
\bibliography{cas-refs}

@article{kerbl2023gaussian,
  title={3D Gaussian Splatting for Real-Time Radiance Field Rendering},
  author={Kerbl, Bernhard and Kopanas, Georgios and Leimk{\"u}hler, Thomas and Drettakis, George},
  journal={ACM Transactions on Graphics},
  volume={42},
  number={4},
  pages={139:1--139:14},
  year={2023},
  publisher={ACM},
  doi={10.1145/3592433}
}

@inproceedings{ma2025shapesplat,
  title={A large-scale dataset of gaussian splats and their self-supervised pretraining},
  author={Ma, Qi and Li, Yue and Ren, Bin and Sebe, Nicu and Konukoglu, Ender and Gevers, Theo and Van Gool, Luc and Paudel, Danda Pani},
  booktitle={2025 International Conference on 3D Vision (3DV)},
  pages={145--155},
  year={2025},
  organization={IEEE},
  doi={10.1109/3DV66043.2025.00019}
}

@inproceedings{li2025scenesplat,
  title={Scenesplat: Gaussian splatting-based scene understanding with vision-language pretraining},
  author={Li, Yue and Ma, Qi and Yang, Runyi and Li, Huapeng and Ma, Mengjiao and Ren, Bin and Popovic, Nikola and Sebe, Nicu and Konukoglu, Ender and Gevers, Theo and Van Gool, Luc and Oswald, Martin R. and Paudel, Danda Pani},
  booktitle={Proceedings of the IEEE/CVF International Conference on Computer Vision},
  pages={4961--4972},
  year={2025}
}

@inproceedings{qin2024langsplat,
  title={Langsplat: 3d language gaussian splatting},
  author={Qin, Minghan and Li, Wanhua and Zhou, Jiawei and Wang, Haoqian and Pfister, Hanspeter},
  booktitle={Proceedings of the IEEE/CVF Conference on Computer Vision and Pattern Recognition},
  pages={20051--20060},
  year={2024}
}

@inproceedings{zhou2024feature,
  title={Feature 3dgs: Supercharging 3d gaussian splatting to enable distilled feature fields},
  author={Zhou, Shijie and Chang, Haoran and Jiang, Sicheng and Fan, Zhiwen and Zhu, Zehao and Xu, Dejia and Chari, Pradyumna and You, Suya and Wang, Zhangyang and Kadambi, Achuta},
  booktitle={Proceedings of the IEEE/CVF Conference on Computer Vision and Pattern Recognition},
  pages={21676--21685},
  year={2024}
}

@inproceedings{guedon2024sugar,
  title={{SuGaR}: Surface-Aligned Gaussian Splatting for Efficient 3D Mesh Reconstruction and High-Quality Mesh Rendering},
  author={Gu{\'e}don, Antoine and Lepetit, Vincent},
  booktitle={Proceedings of the IEEE/CVF Conference on Computer Vision and Pattern Recognition (CVPR)},
  pages={5354--5363},
  year={2024}
}

@inproceedings{lee2024compact,
  title={Compact 3D Gaussian Representation for Radiance Field},
  author={Lee, Joo Chan and Rho, Daniel and Sun, Xiangyu and Ko, Jong Hwan and Park, Eunbyung},
  booktitle={Proceedings of the IEEE/CVF Conference on Computer Vision and Pattern Recognition},
  pages={21719--21728},
  year={2024}
}

@inproceedings{qi2017pointnet,
  title={Pointnet: Deep learning on point sets for 3d classification and segmentation},
  author={Qi, Charles R and Su, Hao and Mo, Kaichun and Guibas, Leonidas J},
  booktitle={Proceedings of the IEEE conference on computer vision and pattern recognition},
  pages={652--660},
  year={2017}
}

@inproceedings{qi2017pointnetplusplus,
  title={PointNet++: Deep Hierarchical Feature Learning on Point Sets in a Metric Space},
  author={Qi, Charles R. and Yi, Li and Su, Hao and Guibas, Leonidas J.},
  booktitle={Advances in Neural Information Processing Systems},
  volume={30},
  year={2017}
}

@article{wang2019dgcnn,
  title={Dynamic graph cnn for learning on point clouds},
  author={Wang, Yue and Sun, Yongbin and Liu, Ziwei and Sarma, Sanjay E and Bronstein, Michael M and Solomon, Justin M},
  journal={ACM Transactions on Graphics (tog)},
  volume={38},
  number={5},
  pages={146:1--146:12},
  year={2019},
  publisher={ACM},
  doi={10.1145/3326362}
}

@inproceedings{xu2018spidercnn,
  title={SpiderCNN: Deep Learning on Point Sets with Parameterized Convolutional Filters},
  author={Xu, Yifan and Fan, Tianqi and Xu, Mengxi and Zeng, Long and Qiao, Yu},
  booktitle={Proceedings of the European Conference on Computer Vision (ECCV)},
  pages={87--102},
  year={2018}
}

@inproceedings{li2018pointcnn,
  title={PointCNN: Convolution On X-Transformed Points},
  author={Li, Yangyan and Bu, Rui and Sun, Mingchao and Wu, Wei and Di, Xinhan and Chen, Baoquan},
  booktitle={Advances in Neural Information Processing Systems},
  volume={31},
  year={2018}
}

@inproceedings{yu2022pointbert,
  title={Point-BERT: Pre-Training 3D Point Cloud Transformers with Masked Point Modeling},
  author={Yu, Xumin and Tang, Lulu and Rao, Yongming and Huang, Tiejun and Zhou, Jie and Lu, Jiwen},
  booktitle={Proceedings of the IEEE/CVF Conference on Computer Vision and Pattern Recognition (CVPR)},
  pages={19313--19322},
  year={2022}
}

@inproceedings{pang2022pointmae,
  title={Masked Autoencoders for Point Cloud Self-Supervised Learning},
  author={Pang, Yatian and Wang, Wenxiao and Tay, Francis E. H. and Liu, Wei and Tian, Yonghong and Yuan, Li},
  booktitle={European Conference on Computer Vision (ECCV)},
  pages={604--621},
  year={2022},
  organization={Springer}
}

@inproceedings{zhang2022pointm2ae,
  title={Point-M2AE: Multi-Scale Masked Autoencoders for Hierarchical Point Cloud Pre-Training},
  author={Zhang, Renrui and Guo, Ziyu and Gao, Peng and Fang, Rongyao and Zhao, Bin and Wang, Dong and Qiao, Yu and Li, Hongsheng},
  booktitle={Advances in Neural Information Processing Systems},
  volume={35},
  pages={27061--27074},
  year={2022}
}

@inproceedings{wang2021occo,
  title={Unsupervised Point Cloud Pre-Training via Occlusion Completion},
  author={Wang, Hanchen and Liu, Qi and Yue, Xiangyu and Lasenby, Joan and Kusner, Matt J.},
  booktitle={Proceedings of the IEEE/CVF International Conference on Computer Vision (ICCV)},
  pages={9782--9792},
  year={2021}
}

@inproceedings{afham2022crosspoint,
  title={Crosspoint: Self-supervised cross-modal contrastive learning for 3d point cloud understanding},
  author={Afham, Mohamed and Dissanayake, Isuru and Dissanayake, Dinithi and Dharmasiri, Amaya and Thilakarathna, Kanchana and Rodrigo, Ranga},
  booktitle={Proceedings of the IEEE/CVF conference on computer vision and pattern recognition},
  pages={9902--9912},
  year={2022}
}

@inproceedings{dong2023act,
  title={Autoencoders as Cross-Modal Teachers: Can Pretrained 2D Image Transformers Help 3D Representation Learning?},
  author={Dong, Runpei and Qi, Zekun and Zhang, Linfeng and Zhang, Junbo and Sun, Jianjian and Ge, Zheng and Yi, Li and Ma, Kaisheng},
  booktitle={International Conference on Learning Representations (ICLR)},
  year={2023}
}

@inproceedings{qi2023recon,
  title={Contrast with Reconstruct: Contrastive 3D Representation Learning Guided by Generative Pretraining},
  author={Qi, Zekun and Dong, Runpei and Fan, Guofan and Ge, Zheng and Zhang, Xiangyu and Ma, Kaisheng and Yi, Li},
  booktitle={Proceedings of the 40th International Conference on Machine Learning},
  pages={28223--28243},
  year={2023},
  volume={202},
  series={Proceedings of Machine Learning Research},
  publisher={PMLR}
}

@inproceedings{xue2023ulip,
  title={Ulip: Learning a unified representation of language, images, and point clouds for 3d understanding},
  author={Xue, Le and Gao, Mingfei and Xing, Chen and Mart{\'\i}n-Mart{\'\i}n, Roberto and Wu, Jiajun and Xiong, Caiming and Xu, Ran and Niebles, Juan Carlos and Savarese, Silvio},
  booktitle={Proceedings of the IEEE/CVF conference on computer vision and pattern recognition},
  pages={1179--1189},
  year={2023}
}

@inproceedings{uy2019scanobjectnn,
  title={Revisiting Point Cloud Classification: A New Benchmark Dataset and Classification Model on Real-World Data},
  author={Uy, Mikaela Angelina and Pham, Quang-Hieu and Hua, Binh-Son and Nguyen, Thanh and Yeung, Sai-Kit},
  booktitle={Proceedings of the IEEE/CVF International Conference on Computer Vision (ICCV)},
  pages={1588--1597},
  year={2019}
}

@inproceedings{wu2015modelnet,
  title={3D ShapeNets: A Deep Representation for Volumetric Shapes},
  author={Wu, Zhirong and Song, Shuran and Khosla, Aditya and Yu, Fisher and Zhang, Linguang and Tang, Xiaoou and Xiao, Jianxiong},
  booktitle={Proceedings of the IEEE Conference on Computer Vision and Pattern Recognition (CVPR)},
  pages={1912--1920},
  year={2015}
}

@article{yi2016shapenetpart,
  title={A Scalable Active Framework for Region Annotation in 3D Shape Collections},
  author={Yi, Li and Kim, Vladimir G. and Ceylan, Duygu and Shen, I-Chao and Yan, Mengyan and Su, Hao and Lu, Cewu and Huang, Qixing and Sheffer, Alla and Guibas, Leonidas},
  journal={ACM Transactions on Graphics},
  volume={35},
  number={6},
  pages={210:1--210:12},
  year={2016}
}

@article{atzmon2018pcnn,
  title={Point Convolutional Neural Networks by Extension Operators},
  author={Atzmon, Matan and Maron, Haggai and Lipman, Yaron},
  journal={ACM Transactions on Graphics},
  volume={37},
  number={4},
  pages={71:1--71:12},
  year={2018}
}

@inproceedings{komarichev2019acnn,
  title={A-CNN: Annularly Convolutional Neural Networks on Point Clouds},
  author={Komarichev, Artem and Zhong, Zichun and Hua, Jing},
  booktitle={Proceedings of the IEEE/CVF Conference on Computer Vision and Pattern Recognition (CVPR)},
  pages={7421--7430},
  year={2019}
}

@inproceedings{yan2020pointasnl,
  title={PointASNL: Robust Point Clouds Processing Using Nonlocal Neural Networks with Adaptive Sampling},
  author={Yan, Xu and Zheng, Chaoda and Li, Zhen and Wang, Sheng and Cui, Shuguang},
  booktitle={Proceedings of the IEEE/CVF Conference on Computer Vision and Pattern Recognition (CVPR)},
  pages={5589--5598},
  year={2020}
}

@inproceedings{radford2021clip,
  title={Learning Transferable Visual Models From Natural Language Supervision},
  author={Radford, Alec and Kim, Jong Wook and Hallacy, Chris and Ramesh, Aditya and Goh, Gabriel and Agarwal, Sandhini and Sastry, Girish and Askell, Amanda and Mishkin, Pamela and Clark, Jack and Krueger, Gretchen and Sutskever, Ilya},
  booktitle={Proceedings of the 38th International Conference on Machine Learning},
  pages={8748--8763},
  year={2021},
  volume={139},
  series={Proceedings of Machine Learning Research},
  publisher={PMLR}
}

@inproceedings{dosovitskiy2021vit,
  title={An Image is Worth 16x16 Words: Transformers for Image Recognition at Scale},
  author={Dosovitskiy, Alexey and Beyer, Lucas and Kolesnikov, Alexander and Weissenborn, Dirk and Zhai, Xiaohua and Unterthiner, Thomas and Dehghani, Mostafa and Minderer, Matthias and Heigold, Georg and Gelly, Sylvain and Uszkoreit, Jakob and Houlsby, Neil},
  booktitle={International Conference on Learning Representations (ICLR)},
  year={2021}
}

@inproceedings{liu2021swin,
  title={Swin Transformer: Hierarchical Vision Transformer Using Shifted Windows},
  author={Liu, Ze and Lin, Yutong and Cao, Yue and Hu, Han and Wei, Yixuan and Zhang, Zheng and Lin, Stephen and Guo, Baining},
  booktitle={Proceedings of the IEEE/CVF International Conference on Computer Vision (ICCV)},
  pages={10012--10022},
  year={2021}
}

@inproceedings{bao2022beit,
  title={{BEiT}: {BERT} Pre-Training of Image Transformers},
  author={Bao, Hangbo and Dong, Li and Piao, Songhao and Wei, Furu},
  booktitle={International Conference on Learning Representations (ICLR)},
  year={2022}
}

@inproceedings{he2016resnet,
  title={Deep Residual Learning for Image Recognition},
  author={He, Kaiming and Zhang, Xiangyu and Ren, Shaoqing and Sun, Jian},
  booktitle={Proceedings of the IEEE Conference on Computer Vision and Pattern Recognition (CVPR)},
  pages={770--778},
  year={2016}
}

@inproceedings{schonberger2016sfm,
  title={Structure-from-Motion Revisited},
  author={Sch{\"o}nberger, Johannes L. and Frahm, Jan-Michael},
  booktitle={Proceedings of the IEEE Conference on Computer Vision and Pattern Recognition (CVPR)},
  pages={4104--4113},
  year={2016}
}

@inproceedings{mildenhall2020nerf,
  title={{NeRF}: Representing Scenes as Neural Radiance Fields for View Synthesis},
  author={Mildenhall, Ben and Srinivasan, Pratul P. and Tancik, Matthew and Barron, Jonathan T. and Ramamoorthi, Ravi and Ng, Ren},
  booktitle={European Conference on Computer Vision (ECCV)},
  pages={405--421},
  year={2020},
  organization={Springer}
}

@article{muller2022instant,
  title={Instant Neural Graphics Primitives with a Multiresolution Hash Encoding},
  author={M{\"u}ller, Thomas and Evans, Alex and Schied, Christoph and Keller, Alexander},
  journal={ACM Transactions on Graphics},
  volume={41},
  number={4},
  pages={102:1--102:15},
  year={2022}
}

@inproceedings{vaswani2017attention,
  title={Attention Is All You Need},
  author={Vaswani, Ashish and Shazeer, Noam and Parmar, Niki and Uszkoreit, Jakob and Jones, Llion and Gomez, Aidan N. and Kaiser, Lukasz and Polosukhin, Illia},
  booktitle={Advances in Neural Information Processing Systems},
  volume={30},
  pages={5998--6008},
  year={2017}
}

@inproceedings{zhu2023pointclipv2,
  title={PointCLIP V2: Prompting CLIP and GPT for Powerful 3D Open-world Learning},
  author={Zhu, Xiangyang and Zhang, Renrui and He, Bowei and Guo, Ziyu and Zeng, Ziyao and Qin, Zipeng and Zhang, Shanghang and Gao, Peng},
  booktitle={Proceedings of the IEEE/CVF International Conference on Computer Vision (ICCV)},
  pages={2639--2650},
  year={2023}
}

@inproceedings{liang2024pointmamba,
  title={PointMamba: A Simple State Space Model for Point Cloud Analysis},
  author={Liang, Dingkang and Zhou, Xin and Xu, Wei and Zhu, Xingkui and Zou, Zhikang and Ye, Xiaoqing and Tan, Xiao and Bai, Xiang},
  booktitle={Advances in Neural Information Processing Systems},
  volume={37},
  pages={32653--32677},
  year={2024},
  doi={10.52202/079017-1026}
}

@inproceedings{han2024mamba3d,
  title={Mamba3D: Enhancing Local Features for 3D Point Cloud Analysis via State Space Model},
  author={Han, Xu and Tang, Yuan and Wang, Zhaoxuan and Li, Xianzhi},
  booktitle={Proceedings of the 32nd ACM International Conference on Multimedia},
  pages={4995--5004},
  year={2024},
  doi={10.1145/3664647.3681173}
}

@inproceedings{zhang2024pointcloudmamba,
  title={Point Cloud Mamba: Point Cloud Learning via State Space Model},
  author={Zhang, Tao and Yuan, Haobo and Qi, Lu and Zhang, Jiangning and Zhou, Qianyu and Ji, Shunping and Yan, Shuicheng and Li, Xiangtai},
  booktitle={Proceedings of the AAAI Conference on Artificial Intelligence},
  volume={39},
  number={10},
  pages={10121--10130},
  year={2025},
  doi={10.1609/aaai.v39i10.33098}
}

@inproceedings{wu2024p2sreslnet,
  title={Point-to-Spike Residual Learning for Energy-Efficient 3D Point Cloud Classification},
  author={Wu, Qiaoyun and Zhang, Quanxiao and Tan, Chunyu and Zhou, Yun and Sun, Changyin},
  booktitle={Proceedings of the AAAI Conference on Artificial Intelligence},
  volume={38},
  number={6},
  pages={6092--6099},
  year={2024},
  doi={10.1609/aaai.v38i6.28425}
}

@article{sun2026align,
  title={Align then Adapt: Rethinking Parameter-Efficient Transfer Learning in 4D Perception},
  author={Sun, Yiding and Zhu, Jihua and Cheng, Haozhe and Lu, Chaoyi and Yang, Zhichuan and Chen, Lin and Wang, Yaonan},
  journal={IEEE Trans. Multimedia},
  year={2026}
}

@article{SUN2026112800,
  title={HyperPoint: Multimodal 3D foundation model in hyperbolic space},
  author={Sun, Yiding and Cheng, Haozhe and Lu, Chaoyi and Li, Zhengqiao and Wu, Minghong and Lu, Huimin and Zhu, Jihua},
  journal={Pattern Recognit.},
  volume={173},
  pages={112800},
  year={2026}
}

@article{li2025pointdico,
  title={PointDico: Contrastive 3D Representation Learning Guided by Diffusion Models},
  author={Li, Pengbo and Sun, Yiding and Cheng, Haozhe},
  journal={arXiv preprint arXiv:2512.08330},
  year={2025}
}

@InProceedings{prcv,
author="Han, Xingguang
and Sun, Yiding
and Lu, Chaoyi",
title="Rethinking Regressor in 3D Gaussian Pretraining",
booktitle="Pattern Recognit. Comput. Vis.",
year="2026",
pages="177--190",
}

@article{zhang2026cmhanet,
  title={CMHANet: A cross-modal hybrid attention network for point cloud registration},
  author={Zhang, Dongxu and Wang, Yingsen and Sun, Yiding and Xu, Haoran and Fan, Peilin and Zhu, Jihua},
  journal={Neurocomputing},
  year={2026},
  publisher={Elsevier}
}

@inproceedings{wang2026pointrft,
  title={PointRFT: Explicit Reinforcement Fine-tuning for Point Cloud Few-shot Learning},
  author={Wang, Yankai and Sun, Yiding and Wang, Qirui and Li, Pengbo and Lu, Chaoyi and Zhang, Dongxu},
  booktitle={IEEE International Conference on Multimedia and Expo (ICME2026)},
  year={2026}
}

@article{guo2026mantis,
  title={Mantis: Mamba-native Tuning is Efficient for 3D Point Cloud Foundation Models},
  author={Guo, Zihao and Zhu, Jihua and Liu, Jian and Mian, Ajmal Saeed},
  journal={arXiv preprint arXiv:2605.03438},
  year={2026},
  doi={10.48550/arXiv.2605.03438}
}

@article{cheng2024multitrusted,
  title={Multi-Trusted Cross-Modal Information Bottleneck for 3D Self-Supervised Representation Learning},
  author={Cheng, Haozhe and Han, Xu and Shi, Pengcheng and Zhu, Jihua and Li, Zhongyu},
  journal={Knowledge-Based Systems},
  volume={283},
  pages={111217},
  year={2024},
  doi={10.1016/j.knosys.2023.111217}
}

@article{cheng2024ptm,
  title={{PTM}: Torus Masking for 3D Representation Learning Guided by Robust and Trusted Teachers},
  author={Cheng, Haozhe and Zhu, Jihua and Hu, Naiwen and Chen, Jinqian and Yan, Wenbiao},
  journal={IEEE Transactions on Circuits and Systems for Video Technology},
  volume={34},
  number={12},
  pages={12158--12170},
  year={2024},
  doi={10.1109/TCSVT.2024.3430904}
}

@inproceedings{hu2024hyperbolic,
  title={Hyperbolic Image-and-Pointcloud Contrastive Learning for 3D Classification},
  author={Hu, Naiwen and Cheng, Haozhe and Xie, Yifan and Shi, Pengcheng and Zhu, Jihua},
  booktitle={2024 IEEE/RSJ International Conference on Intelligent Robots and Systems (IROS)},
  pages={4973--4979},
  year={2024},
  organization={IEEE},
  doi={10.1109/IROS58592.2024.10802543}
}

@article{huang2024joint,
  title={Joint Representation Learning for Text and 3D Point Cloud},
  author={Huang, Rui and Pan, Xuran and Zheng, Henry and Jiang, Haojun and Xie, Zhifeng and Wu, Cheng and Song, Shiji and Huang, Gao},
  journal={Pattern Recognition},
  volume={147},
  pages={110086},
  year={2024},
  doi={10.1016/j.patcog.2023.110086}
}

@article{zhang2025crossmodal,
  title={Cross-Modal Knowledge Transfer for 3D Point Clouds via Graph Offset Prediction},
  author={Zhang, Huang and Yu, Long and Wang, Guoqi and Tian, Shengwei and Yu, Zaiyang and Li, Weijun and Ning, Xin},
  journal={Pattern Recognition},
  volume={162},
  pages={111351},
  year={2025},
  doi={10.1016/j.patcog.2025.111351}
}

@article{zhou2026cpg,
  title={{CPG}: Contrastive Patch-Graph Learning for 3D Point Cloud},
  author={Zhou, Junjie and Song, Yingde and Chiu, Chinwai and Xiong, Yongping and Luo, Yuxin and Song, Siyang},
  journal={Pattern Recognition},
  volume={169},
  pages={111954},
  year={2026},
  doi={10.1016/j.patcog.2025.111954}
}

\end{document}